\documentclass[10pt,journal,compsoc]{IEEEtran}

\usepackage{times}
\usepackage{soul}
\usepackage[utf8]{inputenc}
\usepackage{graphicx}
\usepackage{amsmath}
\usepackage{booktabs}
\usepackage{epsfig}
\usepackage{amssymb}
\usepackage{amsthm}
\usepackage[linesnumbered,boxed,ruled,commentsnumbered]{algorithm2e}
\newtheorem{myDef}{Definition} 
\newtheorem{myTheo}{Theorem}
\newtheorem{myCo}{Corollary}
\usepackage{pgfplots}
\usepackage{tikz}
\usepackage{color}
\usepackage{caption}
\usetikzlibrary{shapes,arrows,calc}

\begin{document}
	\title{LocalDrop: A Hybrid Regularization for Deep Neural Networks}

\author{Ziqing~Lu, 
	Chang~Xu,
	Bo~Du,
	Takashi Ishida,
	Lefei~Zhang,
	and~Masashi Sugiyama
	\IEEEcompsocitemizethanks{\IEEEcompsocthanksitem  Z. Lu, B. Du, and L. Zhang are with the National Engineering Research Center for Multimedia Software, School of Computer Science, Institute of Artificial Intelligence and Hubei Key Laboratory of Multimedia and Network Communication Engineering, Wuhan University, Wuhan 430079, China, 
		\protect\\ E-mail:lzqbob@gmail.com; dubo@whu.edu.cn; zhanglefei@whu.edu.cn.
		\IEEEcompsocthanksitem C. Xu is with the School of Computer Science, Faculty of Engineering, The University of Sydney, Darlington, NSW 2008, Australia 
		\protect\\ E-mail:c.xu@sydney.edu.au.
		\IEEEcompsocthanksitem T. Ishida and M. Sugiyama are with RIKEN Center for Advanced Intelligence Project, and Department of Complexity Science and Engineering, Graduate School of Frontier Sciences, the University of Tokyo. 
		\protect\\ E-mail:ishida@ms.k.u-tokyo.ac.jp; sugi@k.u-tokyo.ac.jp.}
		\thanks {Manuscript received 6 May 2020; revised 16 Dec. 2020; accepted 7 Feb. 2021.}
		\thanks{Corresponding authors: Bo Du and Chang Xu.}
}

\IEEEtitleabstractindextext{
	\begin{abstract}
		In neural networks, developing regularization algorithms to settle overfitting is one of the major study areas. We propose a new approach for the regularization of neural networks by the local Rademacher complexity called LocalDrop. A new regularization function for both fully-connected networks (FCNs) and convolutional neural networks (CNNs), including drop rates and weight matrices, has been developed based on the proposed upper bound of the local Rademacher complexity by the strict mathematical deduction. The analyses of dropout in FCNs and DropBlock in CNNs with keep rate matrices in different layers are also included in the complexity analyses. With the new regularization function, we establish a two-stage procedure to obtain the optimal keep rate matrix and weight matrix to realize the whole training model. Extensive experiments have been conducted to demonstrate the effectiveness of LocalDrop in different models by comparing it with several algorithms and the effects of different hyperparameters on the final performances.
	\end{abstract}
	
}

\maketitle

\section{Introduction}

Neural networks have lately shown impressive performance in sophisticated real-world situations, including image classification \cite{Szegedy2015GoingDW}, object recognition \cite{Yang2015DeepCN} and image captioning \cite{Liu2017MATAM}. Low, middle and high level features are integrated into deep neural networks, which are usually trained in an end-to-end manner. The levels of features can be enriched by stacking more layers (i.e., increasing the network depth) \cite{DBLP:journals/corr/HeZRS15} or widening the layers (i.e., including more filters) \cite{DBLP:journals/corr/ZagoruykoK16}, which then leads to a large number of parameters to be optimized in fully-connected networks (FCNs) and convolutional networks (CNNs). However, given limited training data, it is difficult to figure out appropriate values for such a tremendous volume of variables without overfitting.

Some methods have been developed to enable neural networks to perform well not just on  training data, but also on new inputs \cite{goodfellow2016deep}. Among them, regularization is a major strategy for achieving this goal. The most straightforward solution is to stop the training as soon as the validation performance becomes suboptimal. This tactic is known as early stopping, which is widely used in deep learning regularization. Similarly to other machine learning models, L1 and L2 regularizations are also often adopted to penalize weights in neural networks. Since the seminal work by Hinton \cite{DBLP:journals/corr/abs-1207-0580}, dropout has become a popular technique among deep learning researchers to regularize a broad family of models. It randomly masks out part of the network and trains the ensemble consisting of all sub-networks. Extending the idea of dropout, DropConnect randomly drops weights instead of activations \cite{pmlr-v28-wan13}. Other dropout variants include adaptive dropout \cite{NIPS2013_5032}, variational dropout \cite{Kingma2015VariationalDA} and sparse variational dropout \cite{Molchanov2017VariationalDS}. In convolutional neural networks (CNNs), DropBlock \cite{Ghiasi2018DropBlockAR} extends the idea of dropout to make it adaptable to the structure of CNNs. The original dropout has limit promotions in CNNs because the structure of convolutional networks obliges every unit to correlate with its adjacent units in the feature map. The adjacent units would contain information of any given dropped unit in the feature map. Batch normalization is another popular approach that can evidently improve generalization by reducing the internal covariance shift of feature representations \cite{DBLP:journals/corr/IoffeS15}. Other remarkable regularization techniques include weight decay, spatial shuffling \cite{Hayat2016ASL}, fractional max-pooling \cite{Graham2014FractionalM}, and model averaging \cite{Breiman1996BaggingP}.   

These regularization approaches have been empirically proven to improve the generalization of neural networks and have achieved impressive performance in various applications. There have been several attempts at theoretically explaining the generalization of deep neural networks \cite{Neyshabur2015NormBasedCC,Arora2018StrongerGB,Zheng2019CapacityCO,Arora2019FineGrainedAO} but the link between theory and application is tenuous. Most of the regularization techniques were heuristically developed, only a handful of which were amended and complemented with solid analyses on generalization. For instance, dropout \cite{DBLP:journals/corr/abs-1207-0580} was first sought after thanks to their impressive experimental performance, and later theoretical analyses were conducted in the Bayesian framework \cite{Gal2016DropoutAA} and based on the Rademacher complexity \cite{Gao2015DropoutRC}. However, several recent dropout variants \cite{Molchanov2017VariationalDS,Park2018AdversarialDF,Achille2018InformationDL} have been introduced without theoretical underpinning.

Instead of first heuristically designing a regularization and then seeking for feasible theoretical explanations, we advocate a reverse thinking to distill the regularization technique explicitly from the generalization analyses of neural networks. More specifically, we propose a new regularization algorithm called LocalDrop to regularize neural networks, including FCNs and CNNs, via the local Rademacher complexity of the function class \cite{bartlett2005local}. As the Rademacher complexity provides global estimates of the complexity of a function class, it does not indicate the fact that the algorithm would pick a function that have a small error. Since only a small subset of the function class will be used in practice, we employ the local Rademacher complexity to measure the complexity of neural networks. In FCNs, dropout with varied keep rates across different layers is included in the complexity analyses but DropBlock was replaced in CNNs. For an improved generalization error bound of neural networks, we choose to minimize the local Rademacher complexity of hypotheses, which then provides us a new complexity regularization involving dropout probabilities and weight normalization. A two-stage optimization is developed to efficiently solve neural networks with the new local Rademacher regularization. Experimental results drawn from real-world datasets demonstrate the effectiveness of LocalDrop with rigorous theoretical foundation.

\section{Related Works}

This section is a brief review of regularizations and related theoretical analyses for neural networks.

\subsection{Regularizations for Neural Networks}

Regularization aims to settle overfitting and improve the generalization ability of neural networks. Dropout \cite{DBLP:journals/corr/abs-1207-0580} randomly drops units along with their connections from the neural network, preventing units from overly co-adapting. Then an exponential number of different thinned networks are sampled, in a fashion similar to bagging \cite{Breiman1996BaggingP}. Drawing on the concept of dropout, several studies have been carried out recently. Dropconnect \cite{pmlr-v28-wan13} randomly drops the weights to achieve better performance. Adaptive dropout \cite{NIPS2013_5032} blends a binary belief network with a classic neural network to lower the classification error in practice though theoretically unwarranted. Fast dropout \cite{Wang2013FastDT} replaces actual sampling with an approximation of the Gaussian distribution to accelerate dropout training and to avoid training an additional exponential number of neural networks. Spatial dropout \cite{Tompson2014EfficientOL} randomly drops all the channels from feature maps in CNNs. Variational dropout \cite{Kingma2015VariationalDA} learns dropout rates for better models, which can also be regarded as a generalization of Gaussian dropout. Based on variational dropout, sparse variational dropout \cite{Molchanov2017VariationalDS} uses different dropout rates to obtain a high level sparsity in neural networks. Extending the idea of analyzing dropout from the Bayesian perspective \cite{Gal2016DropoutAA}, concrete dropout \cite{Gal2017ConcreteD} demonstrates the Bayesian generalization of Bernoulli dropout. In the same vein, variational Bayesian dropout \cite{Hron2018VariationalBD} provides the Bayesian generalization of Gaussian dropout. Adversarial dropout \cite{Park2018AdversarialDF} integrates adversarial training and dropout to accomplish state-of-the-art performance in image classification. Information dropout \cite{Achille2018InformationDL} applies information theoretic principles (i.e., Bottleneck principles) to deep learning, including dropout and other strategies. DropBlock \cite{Ghiasi2018DropBlockAR} is a variant of dropout on convolutional neural networks, randomly dropping units in a block region of the feature map. Zoneout \cite{Krueger2017ZoneoutRR} is a variant of dropout on RNNs, randomly preserving units' previous values instead of dropping them. Although these dropout strategies have achieved great success in practice, most of them suffer from the lack of rigorous theoretical support.

\subsection{Theoretical Analyses for Neural Networks}

Since many regularization strategies empirically performed well in various complex applications, the theoretical part of these strategies has been brought to attention recently. Gal and Ghahramani \cite{Gal2016DropoutAA} studied the Bernoulli dropout model from the Bayesian perspective. They justified that dropout can be regarded as a particular situation of Bayesian regularization by combining Bayesian models and deep learning to gauge model uncertainty. Besides, Wan \cite{pmlr-v28-wan13} proposed a generalization bound of dropout via the Rademacher complexity. Gao and Zhou \cite{Gao2015DropoutRC} proved that various dropout strategies can decrease the Rademacher complexity to prevent FCNs from overfitting. Mou \cite{pmlr-v80-mou18a} combined the Rademacher complexity of the hypothesis set and the variance set to determine a generalization upper bound by applying the proposed general framework with random perturbation on parameters. Neyshabur \cite{Neyshabur2015NormBasedCC,Neyshabur2018APA}, Barlett \cite{Bartlett2017SpectrallynormalizedMB}, Golowich \cite{Golowich2018SizeIndependentSC} and Li \cite{Li2018OnTG} obtained different generalization error bounds of the Rademacher complexity in neural networks. Zhai and Wang \cite{zhai2018adaptive} proposed a new regularization algorithm based on the network Rademacher complexity bound by mathematical derivation. More perspectives on generalization of neural networks and regularizations are provided by Neyshabur \cite{Neyshabur2015NormBasedCC}, Arora \cite{Arora2018StrongerGB,Arora2019FineGrainedAO} and Zheng \cite{Zheng2019CapacityCO}.

\section{Complexity Regularization}

Consider a labeled dataset $\mathcal{S}=\{(\textbf{x}_i,\textbf{y}_i)|i\in\{1,2,...,n\},\textbf{x}_i\in\mathbb{R}^d,\textbf{y}_i\in\{0,1\}^k\}$, where $\textbf{x}_i$ is the feature of the $i$-th example, $\textbf{y}_i$ is its corresponding label, and $k$ is the number of classes. Let $k^l$ be the number of neurons in the $l$-th layer of neural networks, where $l\in\{0,1,2,...,L\}$. The first layer (i.e., $l=0$) takes example features $\textbf{x}_i$ as the input, while the last layer (i.e., $l=L$) outputs the prediction $\textbf{y}_i$. Denote $\textbf{W}^l\in \mathbb{R}^{k^{l-1}\times k^{l}}$ as the transformation matrix from the $(l-1)$-th layer to the $l$-th layer. For dropout in FCNs, we denote $\pmb{\theta}^l \in [0,1]^{k^l}$ as the vector of keep rates for the $l$-th layer. In other words, $\pmb{\theta}^l$ represents the possibility of each neuron that is not hidden by dropout in the $l$-th layer. Then we define $\textbf{r}^l\in\{0,1\}^{k^l}$ as a binary vector of the combination of $k^l$ independent Bernoulli dropout random variables (i.e., $\pmb{\theta}^l = \mathbb{E}_{\textbf{r}^l}\textbf{r}^l$). To simplify our notation, we refer $\textbf{W}^{:l}=\{\textbf{W}^1,\textbf{W}^2,...,\textbf{W}^l\}$, $\textbf{r}^{:l}=\{\textbf{r}^0,\textbf{r}^1,...,\textbf{r}^l\}$, $\pmb{\theta}^{:l}=\{\pmb{\theta}^0,\pmb{\theta}^1,...,\pmb{\theta}^l\}$, $\textbf{W}=\textbf{W}^{:L}$, $\textbf{r}=\textbf{r}^{:(L-1)}$ and $\pmb{\theta}=\pmb{\theta}^{:(L-1)}$.
We take ReLU $\phi :\mathbb{R} \rightarrow \mathbb{R}^{+}$ as the activation function. Therefore, we could write the output of the $l$-th layer in vector form in FCNs as
{\small
\begin{equation}\label{FCNf}
f^l(\textbf{x};\textbf{W}^{:l},\textbf{r}^{:(l-1)})=\textbf{W}^l(\textbf{r}^{l-1}\odot\phi(f^{l-1}(\textbf{x};\textbf{W}^{:(l-1)},\textbf{r}^{:(l-2)}))), 
\end{equation}
}\noindent
where $\odot$ reprensents the Hadamard product. As the output of the neural network is a random vector for the sake of the Bernoulli random variables $\textbf{r}$, we take expectation value of $f^L(\textbf{x};\textbf{W},\textbf{r})$ as the deterministic output 
\begin{equation}\label{Ef}
f^L(\textbf{x};\textbf{W},\pmb{\theta})=\mathbb{E}_r[f^L(\textbf{x};\textbf{W},\textbf{r})].
\end{equation}
The final predictions are made through a softmax function, and cross-entropy loss is usually adopted as the optimization objective. According to Wan \cite{pmlr-v28-wan13}, we reformulate the loss function into a logistic function to simplify our generalization analyses as follows
\begin{equation}
\ell(f^L(\textbf{x};\textbf{W},\pmb{\theta }),\textbf{y})=-\sum_{j}y_jlog\dfrac{e^{f^L_j(\textbf{x};\textbf{W},\pmb{\theta})}}{\sum_{j}e^{f^L_j(\textbf{x};\textbf{W},\pmb{\theta})}}.
\end{equation}
We aim to analyze which factors will play important roles in influencing the generalization of the deep neural network, and then incorporate them into the training phase of the neural network for a better generalization.

\subsection{Local Rademacher Complexity of FCNs}

The Rademacher complexity \cite{Bartlett2001RademacherAG} is an effective approach for measuring the complexity of the function class $L$, and it is defined as follows.
\begin{myDef}
	Let $\mathcal{L}$ be a function class mapping from $\textbf{Z}$ to $[a, b]$, and $\mathcal{S}=\{(\textbf{x}_i,\textbf{y}_i)|i\in\{1,2,...,n\}\}$ is a sample of size $n$ in $\textbf{Z}$. Let $\epsilon_1,...,\epsilon_n$ be i.i.d. random variables with $\mathbb{P}[\epsilon_i=1]=\mathbb{P}[\epsilon_i=-1]=\frac{1}{2}$. The Rademacher complexity of function class $\mathcal{L}$ with respect to sample $\mathcal{S}$ is defined as
	\begin{equation}\label{D1}
	R_{\mathcal{S}}(\mathcal{L})=\frac{1}{n}\mathbb{E}_{\epsilon_i}[\sup_{\ell \in \mathcal{L}}\sum_{i=1}^{n}\epsilon_i \ell(\textbf{x}_i,\textbf{y}_i)].
	\end{equation}
\end{myDef}

The Rademacher complexity provides a global estimation of the complexity of a function class. In other words, it does not reflect the fact that the algorithm will likely pick functions that have a small error. Now that only a small subset of the function class will be used, we try to pick out them by adding a restriction (i.e., the variance $\delta$) on function $f$, and then define the empirical local Rademacher complexity.
\begin{myDef}
	Let $\hat{\mathcal{L}}$ be the function class mapping from $\textbf{Z}$ to $[a, b]$, and $\mathcal{\hat{L}}:=\{\hat{\ell}(f): f\in \mathcal{F}\}$, $\mathcal{F}:=\{f|P_n\|f\|_2^2 \leq \delta\}$. $\mathcal{S}=\{(\textbf{x}_i,\textbf{y}_i)|i\in\{1,2,...,n\}\}$ is a sample of size $n$ in $\textbf{Z}$. Then empirical local Rademacher complexity of function class $\mathcal{\hat{L}}$ with respect to the sample $\mathcal{S}$ is defined as
		\begin{equation}\label{D2}
	R_{\mathcal{S}}(\mathcal{\hat{L}})=\frac{1}{n}\mathbb{E}_{\epsilon_i}\big[\sup_{\hat{\ell} \in\mathcal{\hat{L}}}\sum_{i=1}^{n}\epsilon_i \hat{\ell}(\textbf{x}_i)\big],
	\end{equation}
	where $P_n\|f\|_2=\frac{1}{n}\sum_{i=1}^{n}\|f(\textbf{x}_i)\|_2$, and $\|\cdot\|_2$ represents the l2-norm.
\end{myDef}
The local Rademacher complexity provides the complexity estimation of a small subset of a function class. Based on this property, we propose a new upper bound of the local Rademacher complexity of  FCNs with dropout.
\begin{myTheo}
	Assume $\|\textbf{x}\|_\mathrm{F} \leq B$, where $\|\textbf{x}\|_\mathrm{F}$ is the Frobenius norm of the feature of input data and $B \in \mathbb{R}$. Consider $L$ as the number of total layers and $\pmb\theta^l$ as the vector of keep rates in the $l$-th layer. Denote $n$ as the number of examples and $k$ as the number of classes. Consider the singular value decomposition (SVD) of $\textbf{W}^l$ as $\textbf{W}^l=U^l\Sigma ^lV^{lT}$, where $\Sigma^l=\mathrm{diag}(\sigma_1^l,...,\sigma_{\mathrm{rank}(\textbf{W}^l)}^l)$ and $ \sigma_i^l $ is a singular value. Under the condition $\forall h$, $0 \leq  h \leq \mathrm{rank}(\textbf{W}^l)$ and $\frac{1}{n}\sum_{i=1}^{n}\mathbb{E}_{\textbf{r}^{:(l-1)}}\|f^l(\textbf{x}_i;\textbf{W}^{:l},\textbf{r}^{:(l-1)})\|_2^2 \leq \delta^l$, the empirical local Rademacher complexity of the function class $\mathcal{\hat{L}}$ on FCNs is bounded by
	{\small
	\begin{equation}\label{T1}
	\begin{aligned}
	&R_{\mathcal{S}}(\mathcal{\hat{L}}) \leq k\Big[\sqrt{\delta^L}+ \dfrac{B}{n} 2^{L-1} \prod_{i=1}^L (\|\pmb\theta^{i-1}\|_2\sum_{j>h}^{\mathrm{rank}(\textbf{W}^i)}\sigma_j^i)\\
	&+ \sum_{i=2}^{L} 2^{i-1} \sqrt{\delta^{L-i+1}} \prod_{j=1}^{i-1} (\|\pmb\theta^{L-j}\|_2\sum_{j>h}^{\mathrm{rank}(\textbf{W}^{L-j+1})}\sigma_j^{L-j+1})  \Big].
	\end{aligned}
	\end{equation}
	}
\end{myTheo}
The detailed proof of Theorem 1 can be found in the Appendix. Based on the upper bound of the local Rademacher complexity in Theorem 1, a generalization error bound of deep neural networks can be easily derived \cite{ShalevShwartz2014UnderstandingML}. According to Theorem 1, it can be found that increasing the variance $\delta$ will enlarge the upper bound. The upper bound will be tightened when the keep rates $\pmb\theta$ and $\sum_{j>h}^{\mathrm{rank}(\textbf{W})}\sigma_j$ become small. We therefore consider deriving a regularization function in terms of $\pmb\theta$ and $\sum_{j>h}^{\mathrm{rank}(\textbf{W})}\sigma_j$. The local Rademacher complexity of FCNs without dropout can be analyzed in a similar way, and the conclusion is shown in the following corollary. 
\begin{myCo}
	Under the same assumptions as Theorem 1, the empirical local Rademacher complexity of the function class $\mathcal{\hat{L}}$ on FCNs without dropout is bounded by
	\begin{equation}\label{C1}
	\begin{aligned}
	&R_{\mathcal{S}}(\mathcal{\hat{L}}) \leq k\Big[ \sqrt{\delta^L}+ \dfrac{B}{n}2^{L-1} \prod_{i=1}^L (\sum_{j>h}^{\mathrm{rank}(\textbf{W}^i)}\sigma_j^i) \\
	&+ \sum_{i=2}^{L} 2^{i-1} \sqrt{\delta^{L-i+1}} \prod_{j=1}^{i-1} (\sum_{j>h}^{\mathrm{rank}(\textbf{W}^{L-j+1})}\sigma_j^{L-j+1}) \Big].
	\end{aligned}
	\end{equation}
\end{myCo}
In contrast to Theorem 1, the upper bound in Corollary 1 does not include $\pmb\theta$ that is related with dropout. The Rademacher complexity bound of neural networks proposed by Golowich \cite{Golowich2018SizeIndependentSC} could be tighter and independent of the depth of FCNs. But they have additional constraints on loss function and activation function, such as the activation function must be a non element-wise activation function. On the contrary, the activation function we used (ReLU) is an element-wise activation function according to Golowich \cite{Golowich2018SizeIndependentSC}. These constraints are apparently inappropriate in our situation. In addition, we have analyzed the hypotheses complexity of FCNs in a different way by investigating the properties of local hypotheses of FCNs. Moreover, our major aim is to identify the important factors influencing the hypotheses complexity, and incorporate them into the training phase of the neural network. 

\subsection{Complexity Regularization Functions}
According to Theorem 1, the local Rademacher complexity is bounded by a function of keep rates $\pmb\theta$ and variances $\delta$ in FCNs. To improve the hypotheses complexity of FCNs, it is natural to design a regularization function by transforming our proposed upper bound of the local Rademacher complexity. A straight forward regularization function can be written as

{\small
\begin{equation}
\begin{aligned}
&\mathrm{Reg}(\textbf{W},\pmb{\theta})=k\Big[\sqrt{\delta^L}+ \dfrac{B}{n} 2^{L-1} \prod_{i=1}^L (\|\pmb\theta^{i-1}\|_2\sum_{j>h}^{\mathrm{rank}(\textbf{W}^i)}\sigma_j^i)\\
&+ \sum_{i=2}^{L} 2^{i-1} \sqrt{\delta^{L-i+1}} \prod_{j=1}^{i-1} (\|\pmb\theta^{L-j}\|_2\sum_{j>h}^{\mathrm{rank}(\textbf{W}^{L-j+1})}\sigma_j^{L-j+1})  \Big].
\end{aligned}
\end{equation}
}

According to Definition 2, it is difficult to know the exact value of variance $\delta$ of function $f$. For simplicity, we take the variance $\delta$ as a sufficient large constant. In addition, $k$ is the number of classes to predict, $B \in \mathbb{R}$. In this case, by omitting these constants, the regularization function can be rewritten as
\begin{equation}
\begin{aligned}
&\mathrm{Reg}(\textbf{W},\pmb{\theta})=\sum_{i=2}^{L}\prod_{j=1}^{i-1}  (\|\pmb\theta^{L-j}\|_2\sum_{j>h}^{\mathrm{rank}(\textbf{W}^{L-j+1})}\sigma_j^{L-j+1})\\ &+\dfrac{1}{n}\prod_{i=1}^L(\|\pmb\theta^{i-1}\|_2\sum_{j>h}^{\mathrm{rank}(\textbf{W}^i)}\sigma_j^i).
\end{aligned}
 \end{equation}
It can be noticed that both terms in the regularization function involve $\pmb\theta$ and $\sum_{q>h}^{\mathrm{rank}(\textbf{W})}\sigma_q$. The sparseness and the rank of weight matrix in each layer would be influenced by constraining $\pmb\theta$ and $\sum_{q>h}^{\mathrm{rank}(\textbf{W})}\sigma_q$ respectively. For an efficient optimization, we extract these two terms out from the complex regularization function, and achieve 
\begin{equation}
\mathrm{Reg}(\textbf{W},\pmb{\theta})=\sum_{l=1}^{L} \big(\|\pmb{\theta}^{l-1}\|_2\sum_{j>h}^{\mathrm{rank}(\textbf{W}^l)}\sigma_j^l\big).
\end{equation}
Given a training set $\{(x_{i}, y_{i})\}_{i=1}^{n}$,  therefore the objective function of FCNs is defined as
\begin{equation}\label{FCNobj}
\min_{\textbf{W},\pmb{\theta}} \sum_{i=1}^{n}\ell(f(\textbf{x}_i;\textbf{W},\pmb{\theta}), \textbf{y}_i)  + \lambda\sum_{l=1}^{L} \big(\|\pmb{\theta}^{l-1}\|_2\sum_{j>h}^{\mathrm{rank}(\textbf{W}^l)}\sigma_j^l\big). 
\end{equation}
The objective function aims to minimize the differences between the value of $f(\textbf{x}_i)$ and the value of $\textbf{y}_i$. Since overfitting is one of the main problems in deep neural networks, we add a new regularization function into the objective function. This regularization function is derived from the upper bound of the local Rademacher complexity, and it aims to lower the rank of weight matrices $\textbf{W}$ and to penalize $\pmb\theta$.  If there is no dropout in FCNs, the objective function correspondingly becomes
\begin{equation}
\min_{\textbf{W}} \sum_{i=1}^{n}{\ell}(f(\textbf{x}_i;\textbf{W}), \textbf{y}_i) + \sum_{l=1}^{L} \big(\sum_{j>h}^{\mathrm{rank}(\textbf{W}^l)}\sigma_j^l\big).
\end{equation}

\subsection{An Extension to Convolutional Layers}

To extend the complexity regularization to CNNs, we follow the complexity regularization procedure in FCNs. Most of the preliminaries in FCNs are still used in CNNs. Denote $\textbf{W}^l\in \mathbb{R}^{p^l \times q^l}$ as the kernel matrix in the $l$-th layer. Although dropout works well in FCNs, it only has a slight improvement in convolutional neural networks. This is because the structure of convolutional networks makes every unit have correlations with adjacent units in the feature map. If one single unit is dropped in the feature map, the adjacent units also contain some information of the dropped unit. Hence the effect of dropout is much weaker. For the sake of this, DropBlock \cite{Ghiasi2018DropBlockAR} is adopted in our analyses. It randomly drops a block of units in the feature map, which has a better performance than dropout in CNNs. According to DropBlock \cite{Ghiasi2018DropBlockAR}, we calculate the keep rate of each unit in every feature map to achieve mask matrices $\hat{\pmb{\theta}}^l$. Note that in CNNs, the vector $\pmb{\theta}^l$ in FCNs becomes a matrix $\hat{\pmb{\theta}}^l$, which is composed of $\gamma$ and $b$. The detailed matrix $\hat{\pmb\theta}$ can be found in the Appendix. We can correspondingly have $\hat{\pmb{\theta}}^l = \mathbb{E}_{\hat{\textbf{r}}^l}\hat{\textbf{r}}^l$.
Hence, for convolutional neural networks, the output of the $l$-th layer is
\begin{equation}\label{CNNf}
f^l(\textbf{x};\textbf{W}^{:l},\hat{\textbf{r}}^{:(l-1)})=\textbf{W}^l\otimes(\hat{\textbf{r}}^{l-1}\odot\phi(f^{l-1}(\textbf{x};\textbf{W}^{:(l-1)},\hat{\textbf{r}}^{:(l-2)}))),  
\end{equation}
where $\otimes$ means the discrete convolution between matrices. Next, we analyze the local Rademacher complexity of CNNs with DropBlock.

In convolutional networks, we use the entrywise $p$-norm of the matrix instead of the induced $p$-norm of the matrix in mathematics derivation. By doing so, no matter the output of each layer is a vector or matrix (i.e., no matter in FCNs or CNNs), the output with keep rates can be presented in the same way. The entrywise $p$-norm regards an $m\times n$ matrix as an $m\times n$ dimension vector. In other words, the matrix $\textbf{W}$ is converted into vector ($vec(\textbf{W})$) to some extent in the mathematics derivation of convolutional layers. Therefore, the $p$-norms of the output of fully-connected layers and convolutional layers can be regarded as the same in order to remarkably simplify our mathematics derivation. In the whole mathematical part of convolutional networks, all $p$-norms $\|\cdot\|_p$ represent entrywise $p$-norm. The Frobenius form of matrix $\|\cdot\|_\mathrm{F}$ in the condition of theorems and corollaries is a special case of entrywise 2-norm (i.e., $\|\cdot\|_2$). With the idea of DropBlock \cite{Ghiasi2018DropBlockAR}, we can derive the Theorem 2 based on Theorem 1.
{\small
\begin{myTheo}
	Assume $\|\textbf{x}\|_\mathrm{F} \leq B$, where $\|\textbf{x}\|_\mathrm{F}$ is the Frobenius norm of the feature of input data, and $B \in \mathbb{R}$. Denote $\textbf{W}^l\in \mathbb{R}^{p^l \times q^l}$, then assume $p^l \times q^l \leq S^l$, where $S^l \in \mathbb{R}$. Consider $L$ as the number of total layers in CNNs, $\hat{\pmb{\theta}}^l$ as the matrix of keep rates in the $l$-th layer. Denote $n$ as the number of examples, $k$ as the number of classes. Consider the SVD of $\textbf{W}^l$ as $\textbf{W}^l=U^l\Sigma ^lV^{lT}$, where $\Sigma^l=\mathrm{diag}(\sigma_1^l,...,\sigma_{\mathrm{rank}(\textbf{W}^l)}^l)$ and $ \sigma_i^l $ is the singular value. Under the condition $\forall h$, $0 \leq  h \leq \mathrm{rank}(\textbf{W}^l)$, $\frac{1}{n}\sum_{i=1}^{n}\mathbb{E}_{\hat{\textbf{r}}^{:(l-1)}}\|f^l(\textbf{x}_i;\textbf{W}^{:l},\hat{\textbf{r}}^{:(l-1)})\|_2^2 \leq \delta^l$, the empirical local Rademacher complexity of the function class $\mathcal{\hat{L}}$ on CNNs is bounded by
	\begin{equation}\label{T2}
	\begin{aligned}
	&R_{\mathcal{S}}(\mathcal{\hat{L}}) \leq k\big[ \sqrt{\delta^L}+ \dfrac{B}{n}2^{L-1} \prod_{i=1}^L (\sqrt{S^i} \|\hat{\pmb\theta}^{i-1}\|_2 \sum_{j>h}^{\mathrm{rank}(\textbf{W}^i)}\sigma_j^i) \\
	&+ \sum_{i=2}^{L} 2^{i-1} \sqrt{(S\delta)^{L-i+1}} \prod_{j=1}^{i-1} \\
	&(\sqrt{S^{L-j+1}} \|\hat{\pmb\theta}^{L-j}\|_2 \sum_{j>h}^{\mathrm{rank}(\textbf{W}^{L-j+1})}\sigma_j^{L-j+1}) \big],
	\end{aligned}
	\end{equation}
\end{myTheo}
}
The brief proof of Theorem 2 can be found in the Appendix. Theorem 2 further investigates the structure of the network, e.g., $S$ and $\hat{\pmb\theta}$. In mathematics, discrete convolution is composed of addition and multiplication. Hence the entrywise $p$-norm of the convolution of matrices can be transformed to the entrywise $p$-norm of the multiplication of matrices by some inequalities in certain condition. The constant $S$ is also derived from this procedure, converting entrywise 1-norm to entrywise 2-norm. Note that the term $\hat{\pmb\theta}$ in the above upper bound is different from the $\pmb\theta$ in Theorem 1. 

Similarly, the Local Rademacher complexity of CNNs without DropBlock can be analyzed to get the following Corollary.
{\small
\begin{myCo}
	Under the same assumption as Theorem 2, the empirical local Rademacher complexity of the function class $\mathcal{\hat{L}}$ on CNNs without DropBlock is bounded by
	\begin{equation}\label{C2}
	\begin{aligned}
	&R_{\mathcal{S}}(\mathcal{\hat{L}}) \leq k\big[ \sqrt{\delta^L}+ \dfrac{B}{n}2^{L-1} \prod_{i=1}^L (\sqrt{S^i} \sum_{j>h}^{\mathrm{rank}(\textbf{W}^i)}\sigma_j^i) +\\
	& \sum_{i=2}^{L} 2^{i-1} \sqrt{(S\delta)^{L-i+1}} \prod_{j=1}^{i-1} (\sqrt{S^{L-j+1}} \sum_{j>h}^{\mathrm{rank}(\textbf{W}^{L-j+1})}\sigma_j^{L-j+1}) \big],
	\end{aligned}
	\end{equation}
\end{myCo}
}

For convolutional networks with DropBlock, we extract the most influential terms in the upper bound to construct regularization functions. The term $S^l$ in the upper bounds of Theorem 2 and Corollary 2 is also a constant. Whether $\pmb\theta$ or $\hat{\pmb\theta}$ is a vector or matrix does not have any differences, since the 2-norm of vector is same with the entrywise 2-norm of matrix. Thus, we can achieve
\begin{equation}
\min_{\textbf{W},\hat{\pmb\theta}} \sum_{i=1}^{n}\hat{\ell}(f(\textbf{x}_i;\textbf{W},\hat{\pmb\theta}), \textbf{y}_i)  + \lambda\sum_{l=1}^{L} \big(\|\hat{\pmb\theta}^{l-1}\|_2\sum_{j>h}^{\mathrm{rank}(\textbf{W}^l)}\sigma_j^l\big),
\end{equation}
which is same as the objective function in FCNs.
Similarly, if there is no DropBlock in CNNs, the objective function correspondingly becomes
\begin{equation}
\min_{\textbf{W}} \sum_{i=1}^{n}\hat{\ell}(f(\textbf{x}_i;\textbf{W}), \textbf{y}_i) + \sum_{l=1}^{L} \big(\sum_{j>h}^{\mathrm{rank}(\textbf{W}^l)}\sigma_j^l\big).
\end{equation}

\subsection{An Extension to Multiple Channels}

In mainstream CNNs, convolutional feature maps are usually multichannel. In general, each layer's feature map and output have three dimensions, each kernel (weight matrix $\textbf{W}$) has four dimensions. In mathematics derivation, we convert the dimension of output from three to two, and the dimension of kernel from four to two. Each layer's output $\pmb{f}^l$ and kernel $\textbf{W}^l$ are considered as block matrices. Denoting $c^l$ as the channel number of the $l$-th layer's feature map, we have 
\begin{equation}
\pmb{f}^l = \left[
\begin{matrix}
f_1^l & f_2^l & f_3^l & \cdots & f_{c^{l+1}}^l \\
\end{matrix}
\right],
\end{equation}
where $\pmb{f}^l$ means the whole output of the $l$-th layer, and $f_i^l$ (a single block) means one output in the $l$-th layer. Because the number of output is equal to the channel number of the $(l+1)$-th layer's feature map, the subscript of the last block is $ c^{l+1}$. The above equation means that all outputs in the $l$-th layer can be considered as a block matrix $\pmb{f}^l$. According to DropBlock \cite{Ghiasi2018DropBlockAR}, each feature channel is better to have its own DropBlock mask, which implies
\begin{equation}
\hat{\pmb\theta}^{l-1} = \left[
\begin{matrix}
\theta_1^{l-1} & \theta_2^{l-1} & \theta_3^{l-1} & \cdots & \theta_{c^l}^{l-1} \\
\end{matrix}
\right],
\end{equation}
where $\theta_i^{l-1}$ stands for the mask on each output in the $(l-1)$-th layer (i.e., the feature map in the $i$-th channel in the $l$-th layer). For $\pmb\theta = \mathbb{E}_r \textbf{r}$, we can correspondingly get the block matrix $\hat{\textbf{r}}^{l-1}$
\begin{equation}
\hat{\textbf{r}}^{l-1} = \left[
\begin{matrix}
r_1^{l-1} & r_2^{l-1} & r_3^{l-1} & \cdots & r_{c^l}^{l-1} \\
\end{matrix}
\right].
\end{equation}
Based on this idea, we can convert the four dimension's kernel into a two dimension block matrix
\begin{equation}
\textbf{W}^l = \left[
\begin{matrix}
W_{11}^l & W_{12}^l & W_{13}^l & \cdots & W_{1c^{l+1}}^l \\
W_{21}^l & W_{22}^l & W_{23}^l & \cdots & W_{2c^{l+1}}^l \\
\vdots & & & & \vdots\\
W_{c^{l}1}^l & W_{c^{l}2}^l & W_{c^{l}3}^l & \cdots & W_{c^{l}c^{l+1}}^l \\
\end{matrix}
\right],
\end{equation}
where $c^l$ is the channel number of the $l$-th layer, $c^{l+1}$ is the channel number of the $(l+1)$-th layer, the $i$-th row of the block matrix $\textbf{W}^l$ represents all kernels in the $i$-th channel, and the $j$-th colomn of the block matrix $\textbf{W}^l$ represents the $j$-th kernel in all channels. Since the number of kernel in the $l$-th layer is equal to the number of output in the ${l}$-th layer (i.e., the channel number of feature map in the $(l+1)$-th layer), the matrix $\textbf{W}^l$ should have $c^{l+1}$ colomns. Next, $\pmb{f}^l, \hat{\pmb\theta}^l, \hat{\textbf{r}}^l, \textbf{W}^l$ (without subscript) can be regarded as block matrices, and use subscript (i.e., $W_{ij}^l$) to represent the specific block in block matrices.

Based on these block matrices and the Eq. \ref{CNNf}, we have
\begin{equation}\nonumber
\begin{aligned}
\pmb{f}^l = 
[
\sum_{i=1}^{c^{l-1}} \textbf{W}_{i1}^l\otimes \theta_i^{l-1} \phi f_i^{l-1},  \cdots,  \sum_{i=1}^{c^{l-1}} \textbf{W}_{ic^{l+1}}^l\otimes \theta_i^{l-1} \phi f_i^{l-1}
]
\end{aligned}
\end{equation}
where $f_j^l=\sum_{i=1}^{c^{l-1}} \textbf{W}_{ij}^l\otimes \theta_i^{l-1}\phi f_i^{l-1}$. The equation above is a variant of Eq. \ref{CNNf}, presenting the output of the $l$-th layer with multiple channels in convolutional networks. Hence, we can follow Theorem 2 to get a variant of Eq. \ref{T2} in the circumstance of multichannel with DropBlock:
\begin{equation}\nonumber
\begin{aligned}
&R_{\mathcal{S}}(\mathcal{\hat{L}}) \leq k\big[ \dfrac{B}{n} 2^{L-1} \prod_{a=1}^L \sqrt{\pmb{S}^a} \big(\sum_{m=1}^{c^a}(\|\hat{\pmb\theta}_m^{a-1}\|_2 \sum_{r=1}^{c^{a+1}} \sum_{k>h}^{\mathrm{rank}(\textbf{W}_{ij}^a)}\sigma_k^a)\big) \\
&+\sqrt{\delta^L}+ \sum_{a=2}^{L} 2^{a-1} \sqrt{\delta^{L-a+1}} \prod_{j=1}^{a} \sqrt{\pmb{S}^{L-j+1}} \\ &\prod_{j=1}^{a-1}\big(\sum_{m=1}^{c^{L-j+1}}(\|\hat{\pmb\theta}_m^{L-j}\|_2\sum_{r=1}^{c^{a+1}} \sum_{k>h}^{\mathrm{rank}(\textbf{W}_{ij}^a)}\sigma_k^a)\big)  \big],
\end{aligned}
\end{equation}
where $c^l\times c^{l+1}\times p^l\times q^l \leq \pmb{S}^l$ ($W_{ij}^l\in \mathbb{R}^{p^l \times q^l}$, $\textbf{W}^l\in \mathbb{R}^{p^l \times q^l \times c^l\times c^{l+1}}$). The results of these two terms ($\pmb{S}^a$, $\sum_{k>h}^{\mathrm{rank}(\textbf{W}_{ij}^a)}\sigma_k^a$) are consistent with that in Theorem 2. We can follow the process of Section 3.3 to get the regularization function similar to the function 16, which is 
\begin{equation}
\begin{aligned}
\min_{\textbf{W},\hat{\pmb\theta}} &\sum_{i=1}^{n}\hat{\ell}(f(\textbf{x}_i;\textbf{W},\hat{\pmb\theta}), \textbf{y}_i)  + \\
&\lambda\sum_{l=1}^{L} \big[\sum_{m=1}^{c^l} \big(\|\hat{\pmb\theta}_m^{l-1}\|_2 \sum_{j=1}^{c^{l+1}}\sum_{k>h}^{\mathrm{rank}(\textbf{W}_{mj}^l)}\sigma_k^l\big)\big]. 
\end{aligned}
\end{equation}
Network architectures have been considered in the objective function, e.g., the number of channels $c^l$ in the $l$-th layer. Note that the multichannel image convolution is still a 2D convolution, though it looks like 3D convolution. This is because the kernels only slide on the spatial dimensions, rather than the channel dimension. 

\section{Optimization}
Based on the regularization function derived by an upper bound of the local Rademacher complexity, we begin to settle objective functions. In FCNs, there are weight matrices $\textbf{W}$ and keep rate vectors $\pmb{\theta}$ to solve. For optimizing $\pmb\theta$, since both loss function and regularization function contain keep rate vectors $\pmb{\theta}$, block coordinate descent algorithm is used to optimize it \cite{Xu2013ABC}. During the optimization of $\pmb{\theta}$, the expected value of the Bernoulli dropout variables is used to approximate the true value of $f^L(\textbf{x};\textbf{W},\pmb{\theta})$ according to Srivastava \cite{JMLR:v15:srivastava14a}. Because if we compute the true value of $f^L(\textbf{x};\textbf{W},\pmb{\theta})$, it will be extremely time consuming due to the stochasticity of dropout. In other words, the neural network with dropout now works as if it is without dropout. 

Backpropagation is widely used to optimize the weights of FCNs. Nevertheless, we cannot straightforwardly apply backpropagation to solve weight matrices $\textbf{W}$ in Eq. \ref{FCNobj}. Because the proposed new regularization function is a non-smooth convex function, and the non-smooth points may block the backpropagation procedure. Therefore, we propose a two-stage optimization for solving FCNs with the new local Rademacher regularization function. The whole process is demonstrated below. 

Given a fixed keep rate $\pmb\theta$, we first ignore the regularization function and conduct backpropagation based on the classification loss to solve weight matrices $\textbf{W}$. Secondly, we take out the current optimal weight matrix $\widehat{W}^l$ in the $l$-th layer and project it onto the constraint set. 
\begin{equation}
\min_{\textbf{W}^l} \|\textbf{W}^l-\widehat{W}^l\|_2 + C\sum_{j>h}^{\mathrm{rank}(\textbf{W}^l)}\sigma_j^l,
\end{equation}
where $C=\lambda\|\pmb{\theta}^{l-1}\|_2$. Suppose the SVD of $\widehat{W}^l$ is $ \widehat{W}^l= U^l \Sigma^l V^{lT}$. Define an auxiliary matrix $\Sigma^h_C$ based on $\Sigma^l=\mathrm{diag}(\sigma_1^l,...,\sigma_{\mathrm{rank}(\widehat{W}^l)}^l)$ as follows.
{\small
\begin{equation}
\Sigma^h_C={\left[ \begin{array}{cccccc}
	\sigma_1^l&0&0&0&...&0\\
	...&...&...&...&...&...\\
	0&...&\sigma_h^l&0&...&0\\
	0&...&0&\sigma_{h+1}^l-C&...&0\\
	...&...&...&...&...&...\\
	0&0&...&0&\sigma_{\mathrm{rank}(\widehat{W}^l)}^l-C&0\\
	\end{array}
	\right ]}
\end{equation}
}\noindent
According to SVT algorithm \cite{doi:10.1137/080738970}, the solution of Eq. \ref{FCNobj} can be written in a closed form
\begin{equation}
\textbf{W}^l=U^l\Sigma^h_CV^{lT}.
\end{equation}

\begin{algorithm}
	\caption{The Optimization Process in FCNs}
\SetAlgoNoLine
	\KwIn{The labeled dataset $\mathcal{S}$\;\\}
	\KwOut{weight matrix $\textbf{W}$\;\\}
	\BlankLine
	Setting the variance $\delta$\;
	Setting three hyperparameters $\lambda$, $m$ and $h$\;
	Initialize $\textbf{W}$ and $\pmb\theta$, $\textbf{W}:=\textbf{W}^0$, $\pmb\theta:=\pmb\theta^0$\;
	epochNum = 0\;
	\For{epochNum$\leq$epochMaxNum}
	{
		Use backpropagation to update $\textbf{W}$ and $\pmb\theta$\;
		epochNum $:=$ epochNum $+1$\;
		\If {epochNum$\geq$k1 and 
			epochNum mod m == 0}
		{
			$l=1$\;
			\For{$l<L$}
			{
				Get $\Sigma^l$ by $\widehat{W}^l=U^l\Sigma^lV^{lT}$\;
				$C=\lambda\|\pmb{\theta}^{l-1}\|_2$\;
				$\Sigma^h_C=diag(\sigma_1^l,...,\sigma_h^l,\sigma_{h+1}^l-C,...,\sigma_{\mathrm{rank}(\widehat{W}^l)}^l-C)$\;
				$\textbf{W}^l=U^l\Sigma^h_CV^{lT}$\;
				$\widehat{W}^l:=\textbf{W}^l$\;
				$l:=l+1$\;
			}
		}
	}
	\textbf{return} $\textbf{W}$.
\end{algorithm}

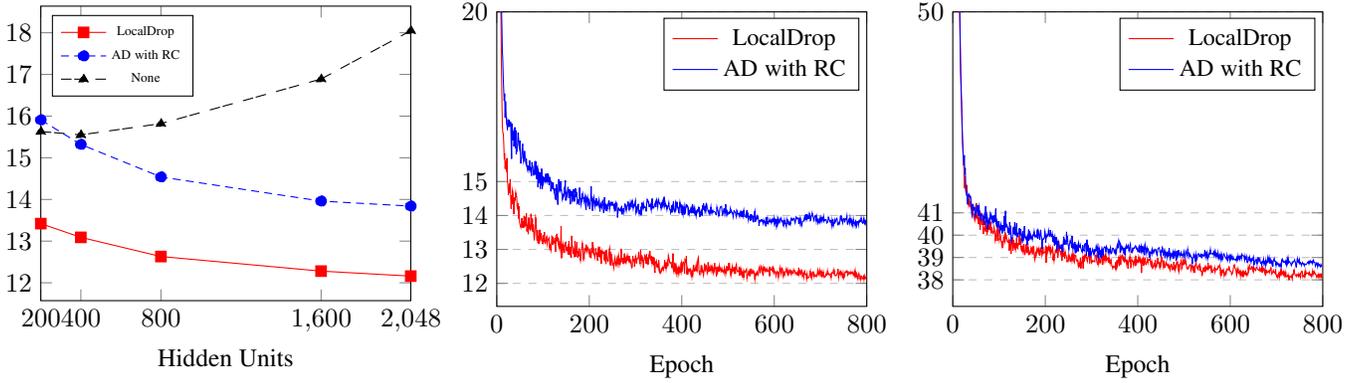
\begin{figure*}
	\centering
	\begin{minipage}{.333\textwidth}
		\begin{tikzpicture}
		\begin{axis}[width=6.5cm, height=5.5cm, xmin=200, xmax=2048, xtick={200,400,800,1600,2048}, xlabel=Hidden Units, ytick={12,13,14,15,16,17,18}, legend pos=north west]
		\addplot[color=red, mark=square*] coordinates {(200,13.42) (400,13.09) (800,12.63) (1600,12.28) (2048,12.16)};
		\addlegendentry{\tiny LocalDrop}
		\addplot[color=blue, mark=oplus*, dash pattern=on 3pt off 2 pt] coordinates {(200,15.91) (400,15.32) (800,14.54) (1600,13.96) (2048,13.84)};
		\addlegendentry{\tiny AD with RC}
		\addplot[color=black, mark=triangle*, dash pattern=on 4pt off 1pt on 4pt off 4pt] coordinates {(200,15.63) (400,15.55) (800,15.82) (1600,16.89) (2048,18.05)};
		\addlegendentry{\tiny None}
		\end{axis}
		\end{tikzpicture}
	\end{minipage}%
	\begin{minipage}{.333\textwidth}
		\begin{tikzpicture}
		\begin{axis}[width=6.5cm, height=5.5cm, xmin=0, xmax=800, ymax=20, ytick={12,13,14,15,20}, xlabel=Epoch, ymajorgrids=true, grid style=dashed,]
		\addplot +[no marks, color=red] table [x=num, y=value] {epoch1.dat};
		\addlegendentry{\small LocalDrop}
		\addplot +[no marks, color=blue] table [x=num, y=value] {epoch2.dat};
		\addlegendentry{\small AD with RC}
		\end{axis}
		\end{tikzpicture}
	\end{minipage}%
	\begin{minipage}{.333\textwidth}
		\begin{tikzpicture}
		\begin{axis}[width=6.5cm, height=5.5cm, xmin=0, xmax=800, ymax=50, ytick={38,39,40,41,50}, xlabel=Epoch, ymajorgrids=true, grid style=dashed,]
		\addplot +[no marks, color=red] table [x=num, y=value] {epoch3.dat};
		\addlegendentry{\small LocalDrop}
		\addplot +[no marks, color=blue] table [x=num, y=value] {epoch4.dat};
		\addlegendentry{\small AD with RC}
		\end{axis}
		\end{tikzpicture}
	\end{minipage}
	\caption{The y-axis of three figures all represent classification error (\%). Left: ($\textbf{a}$) The performance of LocalDrop and adaptive dropout with the Rademacher complexity (i.e., AD with RC) \cite{zhai2018adaptive} to prevent overfitting influenced by the size of two fully-connected layers on CIFAR-10. Middle: ($\textbf{b}$) The performance of LocalDrop and adaptive dropout with the Rademacher complexity \cite{zhai2018adaptive} influenced by epoch number on CIFAR-10. Right: ($\textbf{c}$) The performance of LocalDrop and adaptive dropout with the Rademacher complexity \cite{zhai2018adaptive} influenced by epoch number on CIFAR-100}
\end{figure*}

\pgfplotsset{
	ylabel right/.style={
		after end axis/.append code={
			\node [rotate=90, anchor=north] at (rel axis cs:1.1,0.5) {#1};
		}   
	}
}

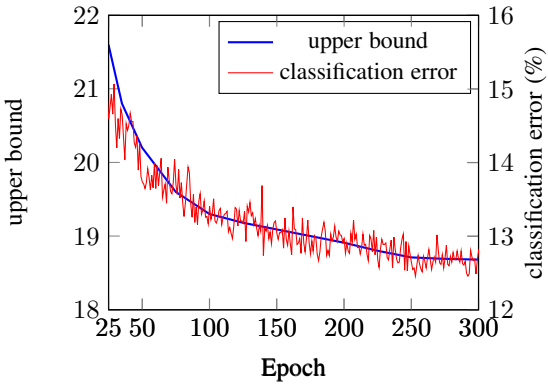
\begin{figure}
	\begin{tikzpicture}
	\begin{axis}[width=6.5cm, height=5.5cm, xmin=25, xmax=300, ymin=18, ymax=22, xtick={25,50,100,150,200,250,300}, ytick={18,19,20,21,22}, xlabel=Epoch, ylabel=upper bound]
	\addplot +[no marks, color=blue, thick] table [x=num, y=value] {epoch5.dat};
	\addlegendentry{\small upper bound}	
	\addplot +[no marks, color=red] table [x=num, y=value] {epoch1.dat};
	\addlegendentry{\small classification error}
	\end{axis}
	\begin{axis}[width=6.5cm, height=5.5cm, xmin=25, xmax=300, ymin=12, ymax=16, xtick={25,50,100,150,200,250,300}, ytick pos=right, ytick={12,13,14,15,16}, xlabel=Epoch, ylabel right=classification error (\%)]
	\addplot +[no marks, color=red] table [x=num, y=value] {epoch1.dat};
	\end{axis}
	\end{tikzpicture}
	\caption{The comparison of the derived upper bound with the classification error (\%) in the optimization process on CIFAR-10 dataset in FCNs. We set $\delta$ as 1 for simplicity. Two y-axes are applied for better visualization of the comparison between the derived upper bound and the actual classification error (\%).}
	\label{Fig:02}
\end{figure}

Then we put the new optimal weight matrix $\textbf{W}^l$ back to the backpropagation procedure by replacing the original weight matrix $\widehat{W}^l$, and continue training the FCNs with forward propagation and backpropagation. For efficiency, the second stage can be conducted once after $m$ ($m={1, 10, 20, 30, \cdots}$) epochs of the first stage. The two-stage optimization method in FCNs is presented below. For FCNs without dropout, the optimization process is basically the same. The only difference is omitting the keep rate vectors $\pmb\theta$ in the functions during the optimization process of FCNs without dropout. The psuedo algorithm is presented above.

For CNNs, we do not exactly follow the procedure in DropBlock \cite{Ghiasi2018DropBlockAR}. Since $\pmb\theta$ is optimized in fully-connected layers, we tend to optimize $\hat{\pmb\theta}$ in convolutional layers. In the matrix $\hat{\pmb\theta}$ (the detailed matrix can be found in the Appendix), there are parameters $b$ and $\gamma$. The estimation of $\gamma$ in DropBlock \cite{Ghiasi2018DropBlockAR} is still used to calculate its value through the following formula
\begin{equation}
\gamma = \frac{d}{b^2}  \frac{t^2}{(t - b+1)^2},
\end{equation}
where $d$ is drop date, $b$ means block size, and $t$ stands for feature size. The term $d$ can be regarded as the probability to drop every unit in traditional dropout. Hence, the above equation explains the relationship between the drop rate in DropBlock (i.e., $\gamma$) and the drop rate in dropout (i.e., $d$) to some extent. The term $b^2$ means every random dropped unit will be expanded to $b^2$ size. The term $(t - b +1)^2$ indicates the valid seed region for dropping seed units. Because we need to make sure that the blocks will not exceed the region of the feature map. Since the valid seeds are all random, there will be some overlaps between dropped blocks. For the sake of this, the equation is an estimation of $\gamma$. The term $b$ is regarded as a hyperparameter, and the term $d$ as a parameter to be optimized. In DropBlock \cite{Ghiasi2018DropBlockAR}, if the initial value of $d$ is big, the performance will be bad. Thus, we set the initial value of $d$ as 0, and use block coordinate descent to achieve the optimal value of $d$, which is the same procedure in fully-connected layers. For optimizing $\textbf{W}^l$, backpropagation is basically the same in both fully-connected layers and convolutional layers.

\begin{table}[!t]
	\centering
	\begin{tabular}{lrr}  
		\toprule
		Model  & CIFAR-10 & CIFAR-100 \\
		\midrule
		Original  &  18.05  &  50.22 \\
		Weight decay      & 17.34  & 46.48   \\
		Early stop   & 16.91  & 45.47  \\
		Dropout & 15.62 & 43.28 \\
		VariationalDropout  &  14.78  &  42.34 \\
		SparseVariationalDropout  &  14.89  &  42.25 \\
		AdaptiveDropout with RC  & 13.79  & 38.57  \\
		LocalDrop in FCNs & 16.67 & 45.29 \\
		LocalDrop without dropout & 16.12  & 45.46  \\
		LocalDrop  & \textbf{12.16}  & \textbf{38.10}  \\
		\bottomrule
	\end{tabular}
	\caption{Classification error (\%) on CIFAR datasets.}
	\label{Tab:01}
\end{table}

\begin{table}[!t]
	\centering
	\begin{tabular}{lrrr}  
		\toprule
		Model  & CIFAR-10 & CIFAR-100 & ILSVRC-2012 \\
		\midrule
		ResNet-50  &  7.3  & 28.9 &  24.5 \\
		+Dropout  & 6.9 & 28.3 & 23.7 \\
		+Cutout      & 6.6  &   27.8  & 24.3 \\
		+SpatialDropout & 6.4 & 27.6 & 23.1 \\
		+DropBlock      & 6.1  & 27.3 & 22.2   \\
		+LocalDrop without DB & 6.2  & 27.5 & 21.9  \\
		+LocalDrop    & \textbf{5.3}  & \textbf{26.2} & \textbf{21.1}  \\
		\midrule
		DenseNet  & 5.0 & 23.6 &  25.1  \\
		+Dropout  & 5.1  &  24.1 &  24.6 \\
		+DropBlock & 4.5 & 23.0 & 24.3 \\
		+LocalDrop  & \textbf{4.2} & \textbf{22.5}  & \textbf{23.8} \\
		\midrule
		RreActResNet  & 4.9 & 22.8 &  21.2  \\
		+Dropout  & 4.9 & 23.3 & 20. 8 \\
		+DropBlock & 4.5 & 22.4 & 20.5 \\
		+LocalDrop  & \textbf{4.3} & \textbf{22.0}  & \textbf{20.2} \\
		\midrule
		DPN-92  & 4.8 & 23.1 &  20.7  \\
		+Dropout & 5.2 & 23.2 & 21.0 \\
		+DropBlock & 4.5 & 22.5 & 20.5 \\
		+LocalDrop  & \textbf{4.3} & \textbf{22.2}  & \textbf{19.9} \\
		\bottomrule
	\end{tabular}
	\caption{Classification error (\%) on CIFAR-10, CIFAR-100 and ILSVRC2012 datasets.}
	\label{Tab:02}
\end{table}

\section{Experiments}
In this section, we conduct experiments to evaluate LocalDrop. We present our experiment settings and illustrate the performance of LocalDrop by comparing it with other representative regularization methods in several different models on CIFAR datasets and ImageNet dataset. Experiments were also done to reveal the properties of LocalDrop. 
\subsection{Experimental Settings}
First, we evaluated the empirical performance of LocalDrop on the CIFAR-10 and CIFAR-100 datasets \cite{Krizhevsky2009LearningML} in Zhai's model \cite{zhai2018adaptive}. These datasets are composed of 60,000 32 $\times$ 32 color images in 10 and 100 different classes respectively. Both of them contain 50,000 training images and 10,000 testing images. According to Srivastava \cite{JMLR:v15:srivastava14a}, all the images were preprocessed by global contrast normalization and ZCA whitening before the training procedure. The neural network architecture design of Zhai \cite{zhai2018adaptive} is adopted for comparison. There are three convolutional layers, each of which is followed by a max-pooling layer. Two fully-connected layers with 2,048 hidden units respectively are followed in the end. LocalDrop was evaluated both in convolutional layers and fully-connected layers.

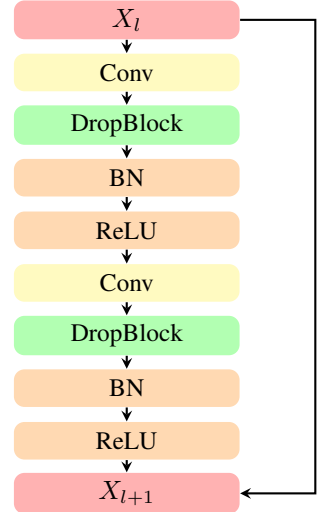
\begin{figure}
\centering
\tikzstyle{start} = [rectangle,rounded corners, minimum width=3cm,minimum height=0.5cm,text centered, fill=red!30]
\tikzstyle{Conv} = [rectangle,rounded corners, minimum width=3cm,minimum height=0.5cm,text centered, fill=yellow!30]
\tikzstyle{Conv2} = [rectangle,rounded corners, minimum width=3cm,minimum height=0.5cm,text centered, fill=yellow!30]
\tikzstyle{BN} = [rectangle,rounded corners, minimum width=3cm,minimum height=0.5cm,text centered, fill=orange!30]
\tikzstyle{BN2} = [rectangle,rounded corners, minimum width=3cm,minimum height=0.5cm,text centered, fill=orange!30]
\tikzstyle{ReLU} = [rectangle,rounded corners, minimum width=3cm,minimum height=0.5cm,text centered, fill=orange!30]
\tikzstyle{ReLU2} = [rectangle,rounded corners, minimum width=3cm,minimum height=0.5cm,text centered, fill=orange!30]
\tikzstyle{DropBlock} = [rectangle,rounded corners, minimum width=3cm,minimum height=0.5cm,text centered, fill=green!30]
\tikzstyle{DropBlock2} = [rectangle,rounded corners, minimum width=3cm,minimum height=0.5cm,text centered, fill=green!30]
\tikzstyle{end} = [rectangle,rounded corners, minimum width=3cm,minimum height=0.5cm,text centered, fill=red!30]
\tikzstyle{arrow} = [thick,->,>=stealth]

\begin{tikzpicture}[node distance=0.7cm]
\node (start) [start,] {$X_l$};
\node (Conv) [Conv,below of=start] {Conv};
\node (DropBlock) [DropBlock,below of=Conv] {DropBlock};
\node (BN) [BN,below of=DropBlock] {BN};
\node (ReLU) [ReLU,below of=BN] {ReLU};
\node (Conv2) [Conv,below of=ReLU] {Conv};
\node (DropBlock2) [DropBlock,below of=Conv2] {DropBlock};
\node (BN2) [BN,below of=DropBlock2] {BN};
\node (ReLU2) [ReLU,below of=BN2] {ReLU};
\node (end) [end,below of=ReLU2] {$X_{l+1}$};

\draw [arrow] (start) -- (Conv);
\draw [arrow] (Conv) -- (DropBlock);
\draw [arrow] (DropBlock) -- (BN);
\draw [arrow] (BN) -- (ReLU);
\draw [arrow] (ReLU) -- (Conv2);
\draw [arrow] (Conv2) -- (DropBlock2);
\draw [arrow] (DropBlock2) -- (BN2);
\draw [arrow] (BN2) -- (ReLU2);
\draw [arrow] (ReLU2) -- (end);
\draw [arrow] (start) -- ($(start.east)+(1,0)$)  |-   (end);
\end{tikzpicture}
\caption{The basic network structure of LocalDrop with DropBlock in ResNet-50}
\label{Fig:03}
\end{figure}

Next, we evaluated the empirical performance of LocalDrop on the CIFAR and ILSVRC2012 datasets in ResNet-50 \cite{He2016DeepRL}. The ILSVRC2012 dataset contains 1.4 million 32 $\times$ 32 color images in 1,000 categories, including 1.2 million training images, 50,000 validation images, and 150,000 testing images. In contrast to DropBlock \cite{Ghiasi2018DropBlockAR}, no data augmentation was used in our experiments. The basic network structure in ResNet is shown in Figure \ref{Fig:03}.

\begin{table}[!t]
	\centering
	\begin{tabular}{lrrrrr}  
		\toprule
		$\lambda$  & 0.01 & 0.03 & 0.1 & 0.3 &  1 \\
		\midrule
		$h$=500,$m$=30  &  12.81 & 12.65 & \textbf{12.56} & 12.75  & 12.95 \\
		$h$=1000,$m$=20  &  12.88 & 12.74 & \textbf{12.61} & 12.82  & 13.02 \\
		$h$=1500,$m$=10  &  12.49 & 12.43 & \textbf{12.38} & 12.51  & 12.82 \\
		$h$=2000,$m$=1  &  12.98 & 12.91 &  \textbf{12.80} & 12.95  & 13.18 \\
		$h$=2000,$m$=10  &  12.34 & 12.27 &  \textbf{12.16} & 12.39  & 12.77 \\
		\bottomrule
	\end{tabular}
	\caption{The influences of $\lambda$ on classification error(\%) of CIFAR-10.}
	\label{Tab:03}
\end{table}

\begin{table*}[!t]
	\centering
	\begin{tabular}{lrrrrrrrrrr}  
		\toprule
		$h$  & 0 & 200 & 500 & 700 &  1000  & 1200 & 1500 & 1700 & 2000 & 2048\\
		\midrule
		$m$=1  &  12.95 & 12.89 & 12.81 & 12.75  & \textbf{12.64} & 12.78 & 12.67 & 12.71 & 12.80 & 13.11 \\
		$m$=10  &  12.42 & 12.55 & 12.75 & 12.66  & 12.78 & 12.74 & 12.38 & 12.30 & \textbf{12.16}$^*$ & 13.11\\
		$m$=20  &  12.68 & 12.75 & 12.72 & 12.63  & 12.61 & 12.47 & \textbf{12.33} & 12.38 & 12.41 & 13.11\\
		$m$=30  &  12.48 & \textbf{12.44} & 12.56 & 12.54  & 12.62 & 12.79 & 12.72 & 12.68  & 12.60 & 13.11\\
		\bottomrule
	\end{tabular}
	\caption{The influences of $h$ and $m$ on classification error(\%) on CIFAR-10 when $\lambda=0.1$. The bold number in each row represents the best performance of that row. The bold number with $^*$ means the best performance of all possible hyperparameter combination choices.}
	\label{Tab:04}
\end{table*}

\begin{table}[!t]
	\centering
	\begin{tabular}{lrr}  
		\toprule
		Model  & CIFAR-10 & CIFAR-100 \\
		\midrule
		Original & 30.33 & 33.43 \\
		Dropout  & 35.57 & 39.22  \\
		AdaptiveDropout with RC      & 38.15  & 42.93   \\
		LocalDrop  & 39.43  & 44.90   \\
		\bottomrule
	\end{tabular}
	\caption{Time comsumption (minutes) on CIFAR datasets every 100 epochs.}
	\label{Tab:05}
\end{table}

\begin{table}[!t]
	\centering
	\begin{tabular}{lrrrrrr}  
		\toprule
		block size & 1 & 3 & 5 & 7  & 9 & 11 \\
		\midrule
		CIFAR-10  &  6.9 & 6.1 & 5.5 & \textbf{5.3}  & 5.8 & 6.6 \\
		CIFAR-100  &  28.3 & 27.0 & 26.6 & \textbf{26.2}  & 26.9 & 27.7\\
		ILSVRC2012  &  23.7 & 22.5 & 21.7 & \textbf{21.1} & 21.4 & 21.9\\
		\bottomrule
	\end{tabular}
	\caption{The influences of block size on classification error(\%) of three datasets.}
	\label{Tab:06}
\end{table}

\begin{table}[!t]
	\centering
	\begin{tabular}{lrrr}  
		\toprule
		Model  & CIFAR-10 & CIFAR-100 & ILSVRC2012 \\
		\midrule
		ResNet-50  &  150  & 167 &  522 \\
		Dropout  & 196 & 212 & 583 \\
		DropBlock      & 204  & 222 & 597   \\
		LocalDrop    & 238  & 267 & 661  \\
		\bottomrule
	\end{tabular}
	\caption{Time comsumption (minutes) on three datasets every 100 epochs.}
	\label{Tab:07}
\end{table}

\subsection{Comparisons with Different Regularizations}
Table \ref{Tab:01} reports the performance of LocalDrop by comparing classification error against other methods on the CIFAR-10 and CIFAR-100 datasets in the first model. The performance of the original network was evaluated in the beginning. Then we realized two common regularization methods (i.e., weight decay and early stop) as the baselines. According to \cite{zhai2018adaptive}, we adopted two variants of dropout (i.e., variational dropout and sparse variational dropout \cite{Kingma2015VariationalDA,Molchanov2017VariationalDS}) which had relatively better performances in various dropout variants. Then the performance of adaptive dropout with the Rademacher complexity \cite{zhai2018adaptive} (i.e., AdaptiveDropout with RC) was evaluated for comparison. Finally, LocalDrop was conducted to demonstrate our algorithm achieving the best performance among these methods. This is mainly because LocalDrop intends to achieve low-rank weight matrices $\textbf{W}$ and to penalize $\pmb\theta$. In addition, two different situations of LocalDrop were evaluated, one only in FCNs (i.e., LocalDrop in FCNs) and the other without dropout (i.e., LocalDrop without dropout) for comparison. It could be found that the regularization in convolutional layers may have bigger effects on the final performance. Hence, we expanded LocalDrop to convolutional layers with DropBlock. Next, we evaluate LocalDrop in ResNet.

Table \ref{Tab:02} shows the performance of LocalDrop comparing to other methods on CIFAR-10, CIFAR-100, and ILSVRC2012 datasets. In comparison with DropBlock \cite{Ghiasi2018DropBlockAR}, these experiments were also conducted in 270 epochs in ResNet-50. Similarly, we first evaluated the performance of the original ResNet-50 \cite{He2016DeepRL} network. Then dropout was regarded as a baseline in ResNet-50, because dropout did not perform well in convolutional layers relative to its performance in FCNs. Next, two algorithms were adopted, Cutout and spatial dropout, which were related to DropBlock \cite{Ghiasi2018DropBlockAR}. Cutout \cite{Devries2017ImprovedRO} is a simplified version of DropBlock, applied to randomly cut out blocks in the input data. DropBlock resembles dropout when $b=1$, and resembles spatial dropout when $b=t$ ($b$ means block size, $t$ means feature size). Therefore, the three algorithms above DropBlock can be regarded as three different special cases of DropBlock. The below three rows present the performances of LocalDrop on several other models, including DenseNet, PreActResNet, DPN. Compared with dropblock, LocalDrop also has lower classification error rates on all datasets. The results in Table \ref{Tab:02} also reflects the low-rank weight matrices will lead to better performance.

Figure 1 ($\textbf{a}$) compares the original network (i.e., None), LocalDrop and adaptive dropout with the Rademacher complexity (i.e., AD with RC) \cite{zhai2018adaptive}. Through analyzing Figure 1 ($\textbf{a}$), it could be found that the performance of the original network (i.e., the black line) deteriorated when the number of hidden units rose in FCNs. But the performance of LocalDrop (i.e., the red line) and adaptive dropout with the Rademacher complexity \cite{zhai2018adaptive} (i.e., the blue line) were still improving when the number of hidden units increased in FCNs. This was because when the size of hidden units became larger, more parameters in weight matrix could lead to overfitting. Rademacher regularization functions were added to both LocalDrop and adaptive dropout with the Rademacher complexity \cite{zhai2018adaptive} both add Rademacher regularization functions to prevent overfitting. In addition, LocalDrop has consistent lower error rates, since it can achieve a low-rank weight matrix.

Figure 1 ($\textbf{b}$) and Figure 1 ($\textbf{c}$) present the changes of the classification error with respect to epoch number on CIFAR-10 and CIFAR-100 datasets. We initialized the learning rate with 0.005 and exponentially decayed it by half every 200 epochs. It can be observed that both LocalDrop and adaptive dropout with the Rademacher complexity \cite{zhai2018adaptive} have relatively stable classification errors between 700 epochs and 800 epochs. Hence we set the maximum of epochs with 800 in the training process. Figure 1 ($\textbf{b}$) and Figure 1 ($\textbf{c}$) again demonstrate that the model performs better when a low-rank weight matrix is achieved. Next, we focus on the analyses of hyperparameter.

Figure \ref{Fig:02} is a visualization of our derived upper bound in section 3.1 changes in the optimization procedure of FCNs. We set $\delta$ as 1 for simplicity. The left y-axis presents the value of our derived upper bound, the right y-axis is the classification error (\%) in empirical experiment. In addition, the derived upper bound was compared with the classification error in the same condition to demonstrate that our bound matches with the trend of classification error. 

Figure \ref{Fig:03} illustrates the basic network structure of LocalDrop in ResNet-50.

\subsection{Hyperparameter Analyses}

For the first model \cite{zhai2018adaptive}, there are three hyperparameters (i.e., $\lambda$, $m$ and $h$) in LocalDrop. In the following, we proceed to analyze the effects of these hyperparameters. 

We begin with $\lambda$. The first four rows in Table \ref{Tab:03} illustrate the effect of $\lambda$ with four rough combinations of $m$ and $h$. It could be observed that for different combinations of $m$ and $h$, the classification error is always the lowest when $\lambda=0.1$. For example, given $h=1500$ and $m=10$, the classification error under $\lambda=1$ is 12.82, which is worse than the error rate under $\lambda=0.1$. Hence, we take 0.1 as the optimal value for hyperparameter $\lambda$. In the last row of Table \ref{Tab:03}, we make sure that initializing parameters with $\{\lambda=0.1, h=2000, m=10\}$ can achieve the best performance on the CIFAR-10 dataset.

Table \ref{Tab:04} shows the influence of hyperparameters $h$ and $m$ on the classification error. Firstly, each row can be considered as an entirety to reveal the effect pattern of $m$. In the condition of using the second stage of our proposed optimization in every epoch (i.e., $m=1$), the result is consistently worse than expected. Because subtracting $C$ from the singular values $\sigma_{h+1}^l, ..., \sigma_{\mathrm{rank}(W^l)}^l$ can reduce the rank of weight matrix $\textbf{W}^l$. If this process is conducted in every epoch, the rank of the weight matrix is too low to elicit good performance. The best result can be achieved when the proposed optimization is conducted in every 10 epochs with $h=2000$. If $m>10$ (i.e., $m={20, 30, ...}$), the performance gradually declined. Because fewer second stages of our proposed optimization procedures had been conducted. The hyperparameter $h$ actually intends to lower the rank of weight matrix $\textbf{W}^l$. A low-rank weight matrix $\textbf{W}^l$ means more correlations between all neurons in the $l$-th layer. However, a lower rank weight matrix $\textbf{W}^l$ does not necessarily lead to better performance. When $h=\mathrm{rank}(\textbf{W}^l)=2048$, no singular value $\sigma_i$ will be subtracted by $C$. So the second stage of our proposed optimization is not used when $h=2048$. The classification error is therefore over $13\%$. After the optimal combination of $m$ and $h$ (i.e., $h=2000, m=10$) is achieved, we apply their values back to Table \ref{Tab:03} to find out whether the assumed best value of $\lambda$ is the optimal value. The processes to find the optimal hyperparameters for the CIFAR-100 dataset was the same as above. The optimal combination of hyperparameters was also $\{\lambda=0.1, h=2000, m=10\}$ on the CIFAR-100 dataset.

Table \ref{Tab:05} shows the time consumption of several algorithms on CIFAR datasets in the first model \cite{zhai2018adaptive}. The experiments in this table were conducted with the optimal value of hyperparameters (i.e., $m=10$). Comparing the last two rows, it only took LocalDrop about 100 seconds more than the algorithm of \cite{zhai2018adaptive} did on every 100 epochs on average. Since the most time-consuming part of LocalDrop is the two-stage optimization procedure in backpropagation, two methods were adopted to speed up this procedure. SVD is time-consuming in the second stage of optimization. According to Cai and Osher \cite{SVT}, we only computed singular values that exceeded a threshold and their associated singular values to accelerate SVD. The partial SVD could be computed efficiently by some Krylov subspace projection methods \cite{Higham1990FastPD}. Next, the second stage of optimization was conducted every $m$ epochs (in this case, $m=10$) instead of every epoch, so the model was much more time efficient. 

For ResNet-50, besides three hyperparameters (i.e., $\lambda$, $m$ and $h$), $b$ (block size) is another hyperparameter to be decided. Firstly, we assumed that the optimal values of $b$ in DropBlock \cite{Ghiasi2018DropBlockAR} (i.e., $b=7$) would perform the best in LocalDrop. Then the above process was repeated to find the optimal values of $\lambda$, $m$ and $h$. The result was the same as the result in fully-connected layers (i.e., $\{\lambda=0.1, h=2, m=10\}$), except for $h$. Since the size of the convolutional kernels in ResNet-50 are $3\times3$ and $1\times1$ (LocalDrop only applied to $3\times3$ kernels), the value of hyperparameters $h$ can only be $1, 2,$ or $3$. When $h=3$, the two-stage optimization did not have any effect (same as when $h=2048$ in the experiment of the first model). It could be found that the best value of hyperparameter $h$ was probably close to the rank of weight matrix $\mathrm{rank}(\textbf{W}^l)$. Thus three hyperparameters with the optimal values were teased out to verify whether the assumption of optimal value of $b$ is true. 

Table \ref{Tab:06} shows the effect of block size on three datasets (CIFAR-10, CIFAR-100, ILSVRC2012). It can be noticed that when the $b$ is 7, the performance is the best in all three datasets. This result corresponds with the results in DropBlock \cite{Ghiasi2018DropBlockAR} in that the performance when $b=9$ was better than that when $b=5$ in the ILSVRC2012 dataset, but this was not the case in CIFAR datasets. The image differences of datasets may be the cause. Therefore, we achieved the optimal combination of hyperparameters (i.e., $\{\lambda=0.1, h=2, m=10, b=7\}$). 

Table \ref{Tab:07} shows the time consumption of several algorithms on three datasets in ResNet-50 every 100 epochs. The experiment was also conducted with the optimal value of hyperparameters $m$ (i.e., $m=10$). Comparing the last two rows, LocalDrop costs about 40 minutes more than DropBlock every 100 epochs in CIFAR datasets, and about 60 minutes more every 100 epochs in the ILSVRC2012 dataset. This is mainly due to the fact that the $d$ (drop rate) is a hyperparameter (linear scheme from 0 to 0.1) in DropBlock, but a parameter, which needs to be optimized, in LocalDrop. In addition, the two-stage optimization process conducted every $m$ epochs also costs extra time (despite the same methods was used in fully-connected layers to speed up).

\section{Conclusion}
In this paper, we have proposed a new regularization algorithm with the local Rademacher complexity for both fully connected networks and convolutional neural networks. In contrast to many regularization algorithms, we provide a rigorous mathematical derivation to achieve the upper bound of the local Rademacher complexity with the analyses of dropout and DropBlock. With the combination of keep rate matrices and weight matrices, we introduce a new regularization function based on the upper bound of the local Rademacher complexity. Therefore a two-stage optimization process is designed to solve neural networks. Experiments on CIFAR-10 dataset, CIFAR-100 dataset, and ILSVRC2012 dataset have proven the superior performance of LocalDrop. Moreover, empirical analyses of several hyperparameters present potential patterns of these parameters. 

\section*{Acknowledgements}
The authors would like to thank the Associate Editor and all anonymous reviewers for their positive support and constructive comments for improving the quality of this paper. This work was supported in part by the National Natural Science Foundation of China under Grants 61822113, National Key Research and Development Program of China under Grants 2018AAA0101100, the Science and Technology Major Project of Hubei Province (Next-Generation AI Technologies) under Grant 2019AEA170, the Natural Science Foundation of Hubei Province under Grants 2018CFA050 and the Supercomputering Center of Wuhan University. C. Xu was supported by ARC DE180101438 and DP210101859. M. Sugiyama was supported by KAKENHI 20H04206. T. Ishida was supported by JST, ACT-X Grant Number JPMJAX2005, Japan.

\section{Appendix}

\subsection{Proof of Theorem 1}

\begin{proof}
	Since the logistic loss function $\ell$ and function $f^L$ are both 1 Lipschitz, based on the Contraction lemma, the local Rademacher complexity of loss function $\mathcal{L}$ can be written as
	\begin{align}
	&R_{\mathcal{S}}(\mathcal{\hat{L}})=\frac{1}{n}\mathbb{E}_{\epsilon_i}\big[\sup_{f^L \in\mathcal{F}}\sum_{i=1}^{n}\epsilon_i \hat{\ell}(f^L(\textbf{x}_i),\textbf{y}_i)\big] \notag \\
	& \leq \frac{k}{n}\mathbb{E}_{\epsilon_i}\big[\sup_{f^L \in\mathcal{F}}\sum_{i=1}^{n}\epsilon_i f^L(\textbf{x}_i)\big]=kR_{\mathcal{S}}(\mathcal{F}^L), \label{kF}
	\end{align}
	where $k$ is the number of classes to predict. To prove the upper bound of $R_{\mathcal{S}}(\mathcal{F}^L)$ in a recursive way, we define a variant of the local Rademacher complexity $\widehat{R}_{\mathcal{S}}(\mathcal{F}^L)$ with 2-norm inside the supremum, which is
	\begin{align}
	&\widehat{R}_{\mathcal{S}}(\mathcal{F}^L)=\frac{1}{n}\mathbb{E}_{\epsilon_i}\big[\sup_{f^L \in\mathcal{F}} \|\sum_{i=1}^{n}\epsilon_i f^L(\textbf{x}_i)\|_2\big]. \label{hatR}
	\end{align}
	We also have
	\begin{align}
	&R_{\mathcal{S}}(\mathcal{F}^L) \leq \widehat{R}_{\mathcal{S}}(\mathcal{F}^L). 
	\end{align}
	Combining Eqs. 1 and 2, we have
	\begin{align}
	&f^l(\textbf{x}_i;\textbf{W}^{:l},\pmb{\theta}^{:(l-1)}) \notag \\
	&=\mathbb{E}_{\textbf{r}^{:l-1}}f^l(\textbf{x}_i;\textbf{W}^{:l},\textbf{r}^{:(l-1)}) \notag \\
	&=\mathbb{E}_{\textbf{r}^{:l-1}}\textbf{W}^l\textbf{r}^{l-1}\odot\phi(f^{l-1}(\textbf{x}_i;\textbf{W}^{:(l-1)},\textbf{r}^{:(l-2)})).
	\end{align}
	To simplify our derivation process, we define
	\begin{align}
	&g^{l-1}(\textbf{x}_i)=\mathbb{E}_{\textbf{r}^{:l-1}}\textbf{r}^{l-1}\odot\phi(f^{l-1}(\textbf{x}_i;\textbf{W}^{:(l-1)},\textbf{r}^{:(l-2)})) \label{glx}, \\
	&g_\epsilon(\textbf{x})=\dfrac{\sum_{i=1}^{n}\epsilon_ig^{l-1}(\textbf{x}_i)}{n} \label{gepsilon}.
	\end{align}
	Hence the variant of the local Rademacher complexity can be rewritten as 
	\begin{align}
	&\widehat{R}_{\mathcal{S}}(\mathcal{F}^l)=\frac{1}{n}\mathbb{E}_{\epsilon_i}\big[\sup_{\textbf{W}^{:l}}\|\sum_{i=1}^{n}\epsilon_if^l(\textbf{x}_i;\textbf{W}^{:l},\pmb{\theta}^{:(l-1)})\|_2\big] \notag\\
	&=\mathbb{E}_{\epsilon_i}\big[\sup_{\textbf{W}^{:l}}\|\frac{1}{n}\sum_{i=1}^{n}\epsilon_if^l(\textbf{x}_i;\textbf{W}^{:l},\pmb{\theta}^{:(l-1)})\|_2\big] \notag \\
	&=\mathbb{E}_{\epsilon_i}\big[\sup_{\textbf{W}^{:l}}\|\frac{1}{n}\sum_{i=1}^{n}\epsilon_i\textbf{W}^lg^{l-1}(\textbf{x}_i)\|_2\big] \notag \\
	&=\mathbb{E}_{\epsilon}\big[\sup_{\textbf{W}^{:l}}\|\textbf{W}^lg_\epsilon(\textbf{x})\|_2\big]. 
	\end{align}
	Considering $\textbf{W}=U\Sigma V$, $\textbf{W}^l$ can be written as
	\begin{align}
	&\textbf{W}^l=\sum_{j=1}^{\mathrm{rank}(\textbf{W}^l)}\sigma_j\textbf{u}_j\textbf{v}_j^T,
	\end{align}
	where $\textbf{u}_j$ and $\textbf{v}_j$ are the column vectors of $U$ and $V$. Then we separate $\textbf{W}^l$ into two parts by $h$.
	$\forall h$, $0 \leq h \leq \mathrm{rank}(\textbf{W}^l)$, we have
	\begin{align}
	&\widehat{R}_{\mathcal{S}}(\mathcal{F}^l)=\mathbb{E}_{\epsilon}\big[\sup_{\textbf{W}^{:l}}\|\sum_{j=1}^{\mathrm{rank}(\textbf{W}^l)}\sigma_j\textbf{u}_j\textbf{v}_j^Tg_\epsilon(\textbf{x})\|_2\big] \notag \\
	&\leq\mathbb{E}_{\epsilon}\big[\sup_{\textbf{W}^{:l}}\|\sum_{j=1}^{h}\sigma_j\textbf{u}_j\textbf{v}_j^Tg_\epsilon(\textbf{x})\|_2\big] \notag \\
	&+\mathbb{E}_{\epsilon}\big[\sup_{\textbf{W}^{:l}}\|\sum_{j>h}^{\mathrm{rank}(\textbf{W}^l)}\sigma_j\textbf{u}_j\textbf{v}_j^Tg_\epsilon(\textbf{x})\|_2\big]. \label{two}
	\end{align}
	After that, we derive the two upper bounds of two terms in the right part of the above inequality separately. Then we combine these two upper bounds with Eqs. \ref{kF} and \ref{hatR} to get the upper bound of the local Rademacher complexity. To simplify the derivation, we denote $F_1$ as the left part of Eq. \ref{two}, $F_2$ as the right part of Eq. \ref{two}. Firstly, we consider $F_1$. Based on Eqs. 1, \ref{glx}, \ref{gepsilon}, and SVD, we have
	\begin{align}
	&F_1=\mathbb{E}_{\epsilon}\big[\sup_{\textbf{W}^{:l}}\|\sum_{j=1}^{h}\sigma_j\textbf{u}_j\textbf{v}_j^T \frac{1}{n}\sum_{i=1}^{n}\epsilon_i g^{l-1}(\textbf{x}_i)\|_2\big] \notag \\
	&\leq\sup_{\textbf{W}^{:l}}\|\sum_{j=1}^{h}\sigma_j\textbf{u}_j\textbf{v}_j^T \frac{1}{n}\sum_{i=1}^{n} g^{l-1}(\textbf{x}_i)\|_2 \notag \\
	&\leq\sup_{\textbf{W}^{:l}}\|\sum_{j=1}^{\mathrm{rank}(\textbf{W}^l)}\sigma_j\textbf{u}_j\textbf{v}_j^T \frac{1}{n}\sum_{i=1}^{n} g^{l-1}(\textbf{x}_i)\|_2 \notag \\
	&\leq \sup_{\textbf{W}^{:l}}(\|\frac{1}{n}\sum_{i=1}^{n}f^l(\textbf{x}_i;\textbf{W}^{:l},\pmb{\theta}^{:(l-1)})\|_2^2)^{\frac{1}{2}} \notag \\
	&\leq \sup_{\textbf{W}^{:l}}(\frac{1}{n}\sum_{i=1}^{n}\|f^l(\textbf{x}_i;\textbf{W}^{:l},\pmb{\theta}^{:(l-1)})\|_2^2)^{\frac{1}{2}} \notag \\
	&\leq \sup_{\textbf{W}^{:l}}(\frac{1}{n}\sum_{i=1}^{n}\mathbb{E}_{\textbf{r}^{:(l-1)}}\|f^l(\textbf{x}_i;\textbf{W}^{:l},\textbf{r}^{:(l-1)})\|_2^2)^{\frac{1}{2}}.
	\end{align}
	Given the restriction of $f^l$ in Theorem 1, we have
	\begin{align}
	&F_1 \leq \sqrt{\delta^l}. \label{F1bound}
	\end{align}
	Secondly, we consider $F_2$. By the equivalence of 1-norm and 2-norm, we have
	\begin{align}
	&F_2 \leq \mathbb{E}_{\epsilon}\big[\sup_{\textbf{W}^{:l}}\sum_{j>h}^{\mathrm{rank}(\textbf{W}^l)}\|\sigma_j^l\textbf{v}_j^Tg_\epsilon(\textbf{x})\|_1\|\textbf{u}_j\|_2\big] \notag \\
	&\leq \mathbb{E}_{\epsilon}\big[\sup_{\textbf{W}^{:l}}\sum_{j>h}^{\mathrm{rank}(\textbf{W}^l)}\|<\sigma_j^l\textbf{v}_j,g_\epsilon(\textbf{x})>\|_1\big] \notag \\
	&\leq \mathbb{E}_{\epsilon}\big[\sup_{\textbf{W}^{:l}}\sum_{j>h}^{\mathrm{rank}(\textbf{W}^l)}\|\sigma_j^l\textbf{v}_j\|_2\|g_\epsilon(\textbf{x})\|_2\big] \notag \\
	&\leq\mathbb{E}_{\epsilon}\big[\sup_{\textbf{W}^{:l}}\|\sum_{j>h}^{\mathrm{rank}(\textbf{W}^l)}\sigma_j^l\|_1\|g_\epsilon(\textbf{x})\|_2\big]. \label{FCNF2}
	\end{align}
	Based on Eqs. \ref{glx} and \ref{gepsilon}, we have
	\begin{align}
	&F_2\leq\sum_{j>h}^{\mathrm{rank}(\textbf{W}^l)}\sigma_j^l\mathbb{E}_{\epsilon}\sup_{\textbf{W}^{:(l-1)}}\|\dfrac{1}{n}\sum_{i=1}^{n}\epsilon_ig^{l-1}(\textbf{x})\|_2 \notag \\
	&\leq \sum_{j>h}^{\mathrm{rank}(\textbf{W}^l)}\sigma_j^l \mathbb{E}_{\epsilon}\sup_{\textbf{W}^{:(l-1)}}\|\dfrac{1}{n}\sum_{i=1}^{n}\epsilon_i(\mathbb{E}_{\textbf{r}^{:l-1}}\textbf{r}^{l-1} \notag \\
	&\odot\phi(f^{l-1}(\textbf{x}_i;\textbf{W}^{:(l-1)},\textbf{r}^{:(l-2)})))\|_2 \notag \\
	&\leq \sum_{j>h}^{\mathrm{rank}(\textbf{W}^l)}\sigma_j^l \dfrac{1}{n}\mathbb{E}_{\epsilon}\sup_{\textbf{W}^{:(l-1)}}\|\sum_{i=1}^{n}\epsilon_i\mathbb{E}_{\textbf{r}^{:l-1}}\textbf{r}^{l-1} \notag \\
	&\odot\phi(f^{l-1}(\textbf{x}_i;\textbf{W}^{:(l-1)},\textbf{r}^{:(l-2)}))\|_2.
	\end{align}
	By Jensen's inequality and Cauchy-Swartz inequality, we have
	\begin{align}
	&F_2\leq \sum_{j>h}^{\mathrm{rank}(\textbf{W}^l)}\sigma_j^l\|\pmb\theta^{l-1}\|_2 \notag \\
	&\dfrac{1}{n}\mathbb{E}_{\epsilon}\sup_{\textbf{W}^{:(l-1)}}\|\sum_{i=1}^{n}\epsilon_i\phi(f^{l-1}(\textbf{x}_i;\textbf{W}^{:(l-1)},\pmb{\theta}^{:(l-2)}))\|_2.
	\end{align}
	Since ReLU $\phi :\mathbb{R} \rightarrow \mathbb{R}^{+}$ is 1-Lipschitz, based on Ledoux-Talagrand contraction we can derive
	\begin{align}
	&F_2\leq \sum_{j>h}^{\mathrm{rank}(\textbf{W}^l)}\sigma_j^l\|\pmb\theta^{l-1}\|_2 \notag \\
	&\dfrac{2}{n}\mathbb{E}_{\epsilon}\sup_{\textbf{W}^{:(l-1)}}\|\sum_{i=1}^{n}\epsilon_if^{l-1}(\textbf{x}_i;\textbf{W}^{:(l-1)},\pmb{\theta}^{:(l-2)})\|_2 \notag \\
	&\leq 2\sum_{j>h}^{\mathrm{rank}(\textbf{W}^l)}\sigma_j^l\|\pmb\theta^{l-1}\|_2\widehat{R}_{\mathcal{S}}(\mathcal{F}^{l-1}). \label{F2bound}
	\end{align}
	Thus we combine Eqs. \ref{F1bound} and \ref{F2bound} to get
	{\small
		\begin{align}
		&\widehat{R}_{\mathcal{S}}(\mathcal{F}^l) \leq F_1+F_2 \leq \sqrt{\delta^l}+2\sum_{j>h}^{\mathrm{rank}(\textbf{W}^l)}\sigma_j^l\|\pmb\theta^{l-1}\|_2\widehat{R}_{\mathcal{S}}(\mathcal{F}^{l-1}). \label{FL}
		\end{align}
	}\noindent
	Consider the fact that
	\begin{align}
	\widehat{R}_{\mathcal{S}}(\mathcal{F}^1) \leq \sqrt{\delta^1}+\sum_{j>h}^{\mathrm{rank}(\textbf{W}^1)}\sigma_j^1\|\pmb\theta^0\|_2\frac{\|\textbf{x}\|_\mathrm{F}}{n}, \label{F1}
	\end{align}
	where $\|\textbf{x}\|_\mathrm{F}$ is the Frobenius norm of the feature of input data. Denote  $\|\textbf{x}\|_\mathrm{F} \leq B$, $B \in \mathbb{R}$. We combine Eqs. \ref{FL}, \ref{F1}, \ref{kF} and \ref{hatR} to achieve
	{\small
		\begin{align}
		&R_{\mathcal{S}}(\mathcal{\hat{L}}) \leq k\Big[\sqrt{\delta^L}+ \dfrac{B}{n} 2^{L-1} \prod_{i=1}^L (\|\pmb\theta^{i-1}\|_2\sum_{j>h}^{\mathrm{rank}(\textbf{W}^i)}\sigma_j^i) \notag \\
		&+ \sum_{i=2}^{L} 2^{i-1} \sqrt{\delta^{L-i+1}} \prod_{j=1}^{i-1} (\|\pmb\theta^{L-j}\|_2\sum_{j>h}^{\mathrm{rank}(\textbf{W}^{L-j+1})}\sigma_j^{L-j+1})  \Big]
		\end{align}
	}\noindent
	which completes the proof.
\end{proof}

\begin{figure*}[!t]
	$$
	\pmb\theta_{drop}^{l-1}=
	\left[
	\begin{matrix}
	\gamma^{l-1}&2\gamma^{l-1}&\cdots&\frac{b+1}{2}\gamma^{l-1}&\cdots&\frac{b+1}{2}\gamma^{l-1}&\cdots&2\gamma^{l-1}&\gamma^{l-1} \\
	2\gamma^{l-1}&4\gamma^{l-1}&\cdots&(b+1)\gamma^{l-1}&\cdots&(b+1)\gamma^{l-1}&\cdots&4\gamma^{l-1}&2\gamma^{l-1}   \\
	\vdots&\vdots& &\vdots& &\vdots& &\vdots&\vdots \\
	\frac{b+1}{2}\gamma^{l-1}&(b+1)\gamma^{l-1}&\cdots& & & &\cdots&(b+1)\gamma^{l-1}&\frac{b+1}{2}\gamma^{l-1} \\
	\vdots&\vdots& & &M_{drop}^{l-1}& & &\vdots&\vdots\\
	\frac{b+1}{2}\gamma^{l-1}&(b+1)\gamma^{l-1}&\cdots& & & &\cdots&(b+1)\gamma^{l-1}&\frac{b+1}{2}\gamma^{l-1} \\
	\vdots&\vdots& &\vdots& &\vdots& &\vdots&\vdots \\
	2\gamma^{l-1}&4\gamma^{l-1}&\cdots&(b+1)\gamma^{l-1}&\cdots&(b+1)\gamma^{l-1}&\cdots&4\gamma^{l-1}&2\gamma^{l-1}   \\
	\gamma^{l-1}&2\gamma^{l-1}&\cdots&\frac{b+1}{2}\gamma^{l-1}&\cdots&\frac{b+1}{2}\gamma^{l-1}&\cdots&2\gamma^{l-1}&\gamma^{l-1} \\
	\end{matrix}
	\right]
	$$ 
\end{figure*}

\begin{figure*}[!t]
	$$
	M_{drop}^{l-1}=
	\left[
	\begin{matrix}
	(\frac{b+1}{2})^2\gamma^{l-1}&\frac{b+1}{2}\frac{b+3}{2}\gamma^{l-1}&\cdots&\frac{b+1}{2}b\gamma^{l-1}&\cdots&\frac{b+1}{2}b\gamma^{l-1}&\cdots&\frac{b+1}{2}\frac{b+3}{2}\gamma^{l-1}&(\frac{b+1}{2})^2\gamma^{l-1}  \\
	\frac{b+1}{2}\frac{b+3}{2}\gamma^{l-1}&(\frac{b+3}{2})^2\gamma^{l-1}&\cdots&\frac{b+3}{2}b\gamma^{l-1}&\cdots&\frac{b+3}{2}b\gamma^{l-1}&\cdots&(\frac{b+3}{2})^2\gamma^{l-1}&\frac{b+1}{2}\frac{b+3}{2}\gamma^{l-1}  \\
	\vdots&\vdots&\ddots&\vdots& &\vdots& &\vdots&\vdots \\
	\frac{b+1}{2}b\gamma^{l-1}&\frac{b+3}{2}b\gamma^{l-1}&\cdots&b^2\gamma^{l-1}&\cdots&b^2\gamma^{l-1}&\cdots&\frac{b+3}{2}b\gamma^{l-1}&\frac{b+1}{2}b\gamma^{l-1}  \\
	\vdots&\vdots& &\vdots& &\vdots& &\vdots&\vdots \\
	\frac{b+1}{2}b\gamma^{l-1}&\frac{b+3}{2}b\gamma^{l-1}&\cdots&b^2\gamma^{l-1}&\cdots&b^2\gamma^{l-1}&\cdots&\frac{b+3}{2}b\gamma^{l-1}&\frac{b+1}{2}b\gamma^{l-1}  \\
	\vdots&\vdots& &\vdots& &\vdots& &\vdots&\vdots \\
	\frac{b+1}{2}\frac{b+3}{2}\gamma^{l-1}&(\frac{b+3}{2})^2\gamma^{l-1}&\cdots&\frac{b+3}{2}b\gamma^{l-1}&\cdots&\frac{b+3}{2}b\gamma^{l-1}&\cdots&(\frac{b+3}{2})^2\gamma^{l-1}&\frac{b+1}{2}\frac{b+3}{2}\gamma^{l-1}  \\
	(\frac{b+1}{2})^2\gamma^{l-1}&\frac{b+1}{2}\frac{b+3}{2}\gamma^{l-1}&\cdots&\frac{b+1}{2}b\gamma^{l-1}&\cdots&\frac{b+1}{2}b\gamma^{l-1}&\cdots&\frac{b+1}{2}\frac{b+3}{2}\gamma^{l-1}&(\frac{b+1}{2})^2\gamma^{l-1}  \\
	\end{matrix}
	\right]
	$$
\end{figure*}

\subsection{The Detailed Matrix}

To better apply DropBlock into mathematics derivation in CNNs, we tend to create the mathematical general form of matrix $\hat{\pmb\theta}^{l-1}$ $(l\in\{1, 2, 3, \cdots, L\})$ based on the idea of DropBlock. The matrix $\hat{\pmb\theta}^{l-1}$ is a mask on the $l$-th layer feature map. In the matrix $\hat{\pmb\theta}^{l-1}$, $b$ represents the $block\ size$, and $\gamma$ is the probability of dropping one unit, which is decided by $drop\ rate$ (traditional dropout rate) and $b$. 

The matrix $\hat{\pmb\theta}^{l-1}$ means the keep probabilities of all units in feature map, but the parameter is $drop\ rate$. Hence the matrix $\hat{\pmb\theta}^{l-1}$ equals $1-\hat{\pmb\theta}_{drop}^{l-1}$. Because the matrix $\hat{\pmb\theta}_{drop}^{l-1}$ is too big to present, we divide it into two parts, $\pmb\theta_{drop}^{l-1}$ and $M_{drop}^{l-1}$. To make sure the block region will be thoroughly contained in feature map, there should be a valid seed region ($M_{drop}^{l-1}$), so the matrix $M_{drop}^{l-1}$ is in the middle of the matrix $\hat{\pmb\theta}^{l-1}$. Denote the matrix $\hat{\pmb\theta}^{l-1} \in \mathbb{R}^{u \times v}$, then the valid seed region $M_{drop}^{l-1}\in \mathbb{R}^{u-(b-1)\times v-(b-1)}$. 
In the matrix $\pmb\theta_{drop}^{l-1}$, the top left term is $\gamma^{l-1}$, because this unit can only be dropped when the top left unit in valid seed region (matrix $M_{drop}^{l-1}$) is dropped. The term on the right of the top left term is $2\gamma^{l-1}$, because this unit can only be dropped when the top left unit or the right unit of the top left unit in valid seed region (matrix $M_{drop}^{l-1}$) is dropped. In this case, we can gradually achieve the drop probabilities of all units in $\pmb\theta_{drop}^{l-1}$. In the matrix $M_{drop}^{l-1}$, the top left unit can be dropped when the unit in the matrix that is expanded by the right $1, 2, \dots, \frac{b+1}{2}$ units and the bottom $1, 2, \dots, \frac{b+1}{2}$ units is dropped. Thus, the drop probability of the top left unit is $(\frac{b+1}{2})^2\gamma^{l-1}$. In the middle of matrix $M_{drop}^{l-1}$, the term $b^2\gamma^{l-1}$ represents this unit can be dropped when the unit in the matrix that is centered by this unit and expanded to the size of $b\times b$ is dropped. Therefore, we can also achieve the drop probabilities of all units in $M_{drop}^{l-1}$. Combine these two matrix $\pmb\theta_{drop}^{l-1}$ and $M_{drop}^{l-1}$, we can get the matrix $\hat{\pmb\theta}_{drop}^{l-1}$, and achieve the mask $\hat{\pmb\theta}^{l-1}$.



\subsection{Brief Proof of Theorem 2}

We only demonstrate the different part in the proof of Theorem 2 comparing to Theorem 1. Other same parts are omitted for simplification. In this section, all $p$-norms $\|\cdot\|_p$ are entrywise $p$-norm (regarding an $m\times n$ matrix as an $m\times n$ dimension vector when calculating the $p$-norm).

\begin{proof}
	
	Similarly with the beginning part of the proof of Theorem 1 (from Eqs. \ref{kF} to \ref{gepsilon}), based on Eq. 13, we can directly have 
	\begin{align}
	&\widehat{R}_{\mathcal{S}}(\mathcal{F}^l)=\frac{1}{n}\mathbb{E}_{\epsilon_i}\big[\sup_{\textbf{W}^{:l}}\|\sum_{i=1}^{n}\epsilon_if^l(\textbf{x}_i;\textbf{W}^{:l},\hat{\pmb\theta}^{:(l-1)})\|_2\big] \notag \\
	&=\mathbb{E}_{\epsilon_i}\big[\sup_{\textbf{W}^{:l}}\|\frac{1}{n}\sum_{i=1}^{n}\epsilon_if^l(\textbf{x}_i;\textbf{W}^{:l},\hat{\pmb\theta}^{:(l-1)})\|_2\big] \notag \\
	&=\mathbb{E}_{\epsilon_i}\big[\sup_{\textbf{W}^{:l}}\|\frac{1}{n}\sum_{i=1}^{n}\epsilon_i\textbf{W}^l\otimes g^{l-1}(\textbf{x}_i)\|_2\big] \notag \\
	&=\mathbb{E}_{\epsilon}\big[\sup_{\textbf{W}^{:l}}\|\textbf{W}^l\otimes g_\epsilon(\textbf{x})\|_2\big].
	\end{align}
	We follow the basic procedure in the proof of Theorem 1, and divide the right part of the above equation into two parts by $h$ to get 
	\begin{align}
	&\widehat{R}_{\mathcal{S}}(\mathcal{F}^l)\leq\mathbb{E}_{\epsilon}\big[\sup_{\textbf{W}^{:l}}\|\sum_{j=1}^{h}\sigma_j\textbf{u}_j\textbf{v}_j^T \otimes g_\epsilon(\textbf{x})\|_2\big] \notag \\
	&+\mathbb{E}_{\epsilon}\big[\sup_{\textbf{W}^{:l}}\|\sum_{j>h}^{\mathrm{rank}(W^l)}\sigma_j\textbf{u}_j\textbf{v}_j^T \otimes g_\epsilon(\textbf{x})\|_2)\big]. \label{CNNtwo}
	\end{align}
	Denote $\hat{F}_1$ as the left part of Eq. \ref{CNNtwo}, and $\hat{F}_2$ as the right part of Eq. \ref{CNNtwo}. Firstly, we consider $\hat{F}_1$. According to SVD and properties of entrywise 2-norm, we have
	{\small
		\begin{align}
		&\hat{F}_1=\mathbb{E}_{\epsilon}\big[\sup_{\textbf{W}^{:l}}\|\sum_{j=1}^{h}\sigma_j\textbf{u}_j\textbf{v}_j^T \otimes \frac{1}{n}\sum_{i=1}^{n}\epsilon_i g^{l-1}(\textbf{x}_i)\|_2\big] \notag \\
		&\leq\sup_{\textbf{W}^{:l}}\|\sum_{j=1}^{\mathrm{rank}(\textbf{W}^l)}\sigma_j\textbf{u}_j\textbf{v}_j^T \otimes \frac{1}{n}\sum_{i=1}^{n}g^{l-1}(\textbf{x}_i)\|_2 \notag \\
		&\leq\sup_{\textbf{W}^{:l}} (\|\textbf{W}^l \otimes \frac{1}{n}\sum_{i=1}^{n}g^{l-1}(\textbf{x}_i)\|_2^2)^\frac{1}{2} \notag \\
		&\leq \sup_{\textbf{W}^{:l}}(\frac{1}{n}\sum_{i=1}^{n}\|f^l(\textbf{x}_i;\textbf{W}^{:l},\pmb{\theta}^{:(l-1)})\|_2^2)^{\frac{1}{2}} \notag \\
		&\leq \sqrt{\delta^l}.  \label{hatF1}
		\end{align}
	}\noindent
	Secondly, we consider $\hat{F}_2$. Our goal is to transform the $\hat{F}_2$ to functions that do not have discrete convolution $\otimes$. According to the Young's convolution inequality in \cite{maths}, regarding the entrywise $p$-norm of matrix as the $p$-norm of vector, the entrywise 2-norm of the discrete convolution of $\sum_{j>h}^{\mathrm{rank}(\textbf{W}^l)}\sigma_j\textbf{u}_j\textbf{v}_j^T$ and $g_\epsilon(\textbf{x})$ can be transform to the product of entrywise 1-norm of $\sum_{j>h}^{\mathrm{rank}(\textbf{W}^l)}\sigma_j\textbf{u}_j\textbf{v}_j^T$ and entrywise 2-norm of $g_\epsilon(\textbf{x})$. Hence, we have
	\begin{align}
	&\hat{F}_2=\mathbb{E}_{\epsilon}\big[\sup_{\textbf{W}^{:l}}\|\sum_{j>h}^{\mathrm{rank}(\textbf{W}^l)}\sigma_j^l \textbf{u}_j\textbf{v}_j^T\|_1\|g_\epsilon(\textbf{x})\|_2\big].
	\end{align}
	The properties of entrywise $p$-norm of matrix is same with the $p$-norm of vector. Based on the relations between 1-norm and 2-norm in vector space, we have 
	\begin{align}
	&\hat{F}_2=\mathbb{E}_{\epsilon}\big[\sup_{\textbf{W}^{:l}}\sqrt{S^l}\|\sum_{j>h}^{\mathrm{rank}(\textbf{W}^l)}\sigma_j^l \textbf{u}_j\textbf{v}_j^T\|_2\|g_\epsilon(\textbf{x})\|_2\big].
	\end{align}
	where $S^l \in \mathbb{R}, l \in \{1, 2, 3, \cdots, L\}$ ($\textbf{W}^l\in \mathbb{R}^{p^l \times q^l}, p^l \times q^l \leq S^l$). According to \cite{Kalman1996ASV}, based on the properties of entrywise 2-norm, we can directly have
	\begin{align}
	&\hat{F}_2 \leq \mathbb{E}_{\epsilon}\big[\sup_{\textbf{W}^{:l}}\sqrt{S^l}\|\sum_{j>h}^{\mathrm{rank}(\textbf{W}^l)}\sigma_j^l\|_2\|g_\epsilon(\textbf{x})\|_2\big].
	\end{align}
	Based on the relations between 1-norm and 2-norm in vector space, we have 
	\begin{align}
	&\hat{F}_2 \leq \mathbb{E}_{\epsilon}\big[\sup_{\textbf{W}^{:l}}\sqrt{S^l}\|\sum_{j>h}^{\mathrm{rank}(\textbf{W}^l)}\sigma_j^l\|_1\|g_\epsilon(\textbf{x})\|_2\big].
	\end{align}
	The above inequality is the same with Eq. \ref{FCNF2}, except for a constant $\sqrt{S^l}$. Hence, we can follow the exact procedure from Eq. \ref{FCNF2} to Eq. \ref{F2bound} in the proof of Theorem 1 to achieve
	\begin{align}
	&\hat{F}_2\leq 2 \sqrt{S^l}\sum_{j>h}^{\mathrm{rank}(\textbf{W}^l)}\sigma_j^l\|\pmb\theta^{l-1}\|_2\widehat{R}_{\mathcal{S}}(\mathcal{F}^{l-1}). \label{hatF2}
	\end{align}
	Similarly with the last step in the proof of Theorem 1, we combine Eqs. \ref{hatF1} and \ref{hatF2} to achieve 
	\begin{align}
	&R_{\mathcal{S}}(\mathcal{\hat{L}}) \leq k\big[ \sqrt{\delta^L}+ \dfrac{B}{n}2^{L-1} \prod_{i=1}^L (\sqrt{S^i} \|\hat{\pmb\theta}^{i-1}\|_2 \sum_{j>h}^{\mathrm{rank}(\textbf{W}^i)}\sigma_j^i) \notag \\
	&+ \sum_{i=2}^{L} 2^{i-1} \sqrt{(S\delta)^{L-i+1}} \prod_{j=1}^{i-1} \notag \\
	&(\sqrt{S^{L-j+1}} \|\hat{\pmb\theta}^{L-j}\|_2 \sum_{j>h}^{\mathrm{rank}(\textbf{W}^{L-j+1})}\sigma_j^{L-j+1}) \big],
	\end{align}
	which completes the proof.
\end{proof}

\bibliographystyle{IEEEtran}
\bibliography{TPAMI}

\end{document}